\documentclass{article}

\def\isaccepted{1}

\usepackage{microtype}
\usepackage{graphicx}
\usepackage{subcaption}
\usepackage{booktabs} 

\usepackage{icml2026}

\usepackage{amsmath}
\usepackage{amssymb}
\usepackage{mathtools}
\usepackage{amsthm}
\usepackage{url}
\usepackage{dsfont}
\usepackage{graphicx}
\usepackage{multirow}
\usepackage{booktabs}
\usepackage{amsmath}
\usepackage{amssymb}
\usepackage{xcolor}
\usepackage[table]{xcolor}
\usepackage{pifont}
\usepackage{caption}
\usepackage{subcaption}
\usepackage{wrapfig}

\definecolor{purple}{rgb}{0.56,0.27,0.68}
\definecolor{newred}{rgb}{0.95,0.4,0.4}
\definecolor{purered}{rgb}{1,0,0}
\definecolor{blue}{rgb}{0.4,0.4,0.95}
\definecolor{darkblue}{rgb}{0,0,0.8}
\definecolor{grey}{rgb}{0.6,0.6,0.6}
\definecolor{col1}{RGB}{232, 161, 148}
\definecolor{col2}{RGB}{148, 187, 232}
\definecolor{col3}{RGB}{206, 239, 255}
\definecolor{lightgrey}{rgb}{0.85,0.85,0.85}
\definecolor{lightlightgrey}{rgb}{0.9,0.9,0.9}
\definecolor{verylightBG}{rgb}{0.9,0.99,0.99}
\definecolor{darkgreen}{rgb}{0.3, 0.75, 0.3}
\definecolor{orange}{rgb}{1.0,0.65,0.1}
\definecolor{darkorange}{rgb}{1.0,0.549,0.0}

\definecolor{cvprblue}{rgb}{0.21,0.49,0.74}
\usepackage[colorlinks=true,      
            linkcolor=darkorange,       
            citecolor=cvprblue,      
            urlcolor=magenta]     
            {hyperref}

\usepackage[capitalize,noabbrev]{cleveref}

\theoremstyle{plain}

\theoremstyle{definition}

\theoremstyle{remark}

\usepackage[textsize=tiny]{todonotes}

\icmltitlerunning{ReBA-Pred-Net}

\begin{document}

\twocolumn[
  \icmltitle{ReBA-Pred-Net: Weakly-Supervised Regional Brain Age Prediction on MRI}



  \icmlsetsymbol{equal}{*}

  \begin{icmlauthorlist}
    \icmlauthor{Shuai Shao}{addressone}
    \icmlauthor{Yan Wang}{addressfour}
    \icmlauthor{Shu Jiang}{addresstwo}
    \icmlauthor{Shiyuan Zhao}{addressthree}
    \icmlauthor{Xinzhe Luo}{addressone}
    \icmlauthor{Di Yang}{addressone}\\
    \icmlauthor{Jiangtao Wang}{addressone}    
    \icmlauthor{Yutong Bai}{addressfive}
    \icmlauthor{Jianguo Zhang}{addressfive}
  \end{icmlauthorlist}

  \icmlaffiliation{addressone}{University of Science and Technology of China, Hefei, China}
  \icmlaffiliation{addresstwo}{China University of Petroleum (East China), Qingdao, China}
  \icmlaffiliation{addressthree}{Northwestern Polytechnical University, Xian, China}
  \icmlaffiliation{addressfour}{Beijing Jiaotong University, Beijing, China}
  \icmlaffiliation{addressfive}{Beijing Tiantan Hospital, Beijing, China}

  \icmlcorrespondingauthor{Yan Wang}{wangyan9509@gmail.com}
  \icmlcorrespondingauthor{Jiangtao Wang}{wangjiangtao@ustc.edu.cn}
  \icmlcorrespondingauthor{Jianguo Zhang}{zjguo73@126.com}

  \icmlkeywords{Machine Learning, ICML}

  \vskip 0.3in
]



\printAffiliationsAndNotice{\icmlEqualContribution}

\begin{abstract}

Brain age has become a prominent biomarker of brain health. Yet most prior work targets whole brain age (WBA), a coarse paradigm that struggles to support tasks such as disease characterization and research on development and aging patterns, because relevant changes are typically region-selective rather than brain-wide. 
Therefore, robust regional brain age (ReBA) estimation is critical, yet a widely generalizable model has yet to be established.
In this paper, we propose the \textbf{Regional Brain Age Prediction Network (ReBA-Pred-Net)}, a Teacher-Student framework designed for fine-grained brain age estimation. 
The Teacher produces soft ReBA to guide the Student to yield reliable ReBA estimates with a clinical-prior consistency constraint (regions within the same function should change similarly).
For rigorous evaluation, we introduce two indirect metrics: \textbf{Healthy Control Similarity (HCS)}, which assesses statistical consistency by testing whether regional brain-age-gap (ReBA minus chronological age) distributions align between training and unseen HC; and \textbf{Neuro Disease Correlation (NDC)}, which assesses factual consistency by checking whether clinically confirmed patients show elevated brain-age-gap in disease-associated regions.
Experiments across multiple backbones demonstrate the statistical and factual validity of our method.

\end{abstract}

\section{Introduction}
\label{sec: intro}


As individuals age, the human brain undergoes progressive senescence, such as reductions in gray-matter volume, degeneration of white-matter microstructure, and remodeling of functional connectivity. 
To quantitatively characterize these age-related alterations, researchers have introduced the \textit{Brain Age} representation that maps neuroimaging features (\textit{e.g.}, Magnetic Resonance Imaging, MRI) to chronological age in healthy control (HC) \cite{wang2025full,lee2022deep,bethlehem2022brain}.
The difference between brain age and chronological age (Brain Age Gap) indicates whether a brain appears “older” or “younger” than expected and has become an important composite biomarker for assessing brain health.


\begin{figure*}[t]
\centering
\includegraphics[width=1.0\textwidth]{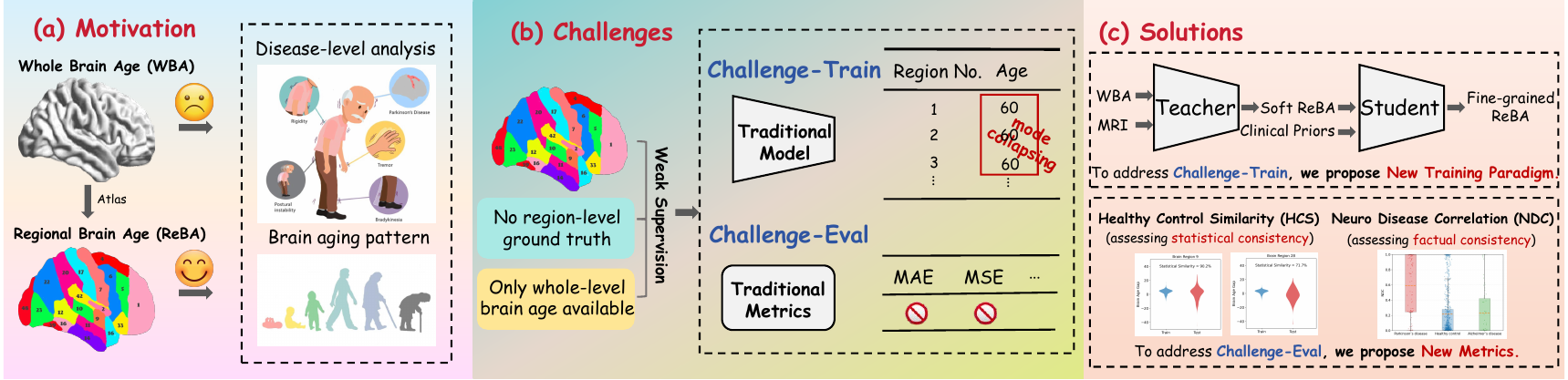} 
\vspace{-5mm}
\caption{ \small
\textbf{(a) Motivation}.
The coarse WBA cannot adequately support disease-level analyses or studies of aging. Parcellating the brain with established functional atlases and modeling ReBA has the potential to solve these limitations and to strengthen brain-health research.
\textbf{(b) Challenges}.
ReBA lacks region-level ground truth. In training, HC chronological age is the only available weak signal, risking mode collapsing and obscuring true regional differences; in evaluation, the absence of regional truth prevents region-wise supervision, making standard metrics (\textit{e.g.}, MAE/MSE) inapplicable.
\textbf{(c) Solutions}.
We propose ReBA-Pred-Net with Teacher-Student framework to address the training problem, and introduce two evaluation metrics, HCS and NDC, to assess statistical consistency and factual consistency.
}
\label{fig: introduction}
\end{figure*}

Conventional brain age prediction typically treats the whole-brain as the analytical unit, yet such a coarse-grained paradigm struggles to support many neuroscientific tasks:
\textbf{(i)} Brain aging patterns. Brain maturation and senescence are heterogeneous. 
Brain regions differ in maturation timelines, peak volumes, and atrophy rates
(\textit{e.g.}, sensorimotor-related brain regions often peak earlier than prefrontal brain regions). 
Whole Brain Age (WBA) cannot reveal these asynchronous, region-specific timelines and rate differences.
\textbf{(ii)} Disease-level analysis. Pathological changes in neurological disorders are region-selective rather than brain-wide. \textit{e.g.}, Parkinson’s disease (PD) primarily involves motor-related regions, whereas Alzheimer’s disease (AD) is closely linked to memory-related structures.
However, WBA cannot localize affected regions, making it difficult to support pathological interpretation and clinical decision-making. 
Therefore, \textit{moving to a finer spatial granularity with Regional Brain Age (ReBA) prediction}, is vital for both neuroscience research and clinical applications (see Fig.~\ref{fig: introduction} (a)).

Recently, WBA prediction is relatively mature, whereas ReBA estimation remains nascent.
Existing approaches are either purely conceptual~\cite{bethlehem2022brain,kalc2024brainage} or rely on pre-extracted morphological features (e.g., volume, thickness)~\cite{riccardi2025distinct,lee2022regional}, which suffer from spatial information loss and poor generalizability due to scanner and pipeline sensitivities. We aim to design a feature-engineering-free deep-learning-based framework operating directly on raw 3D MRI.

However, realizing this goal exposes a \textit{fundamental bottleneck: the absence of observable ground-truth labels for ReBA}.
This limitation means our only viable option is to use HC’ chronological age as weak supervision to predict ReBA, which unavoidably brings substantial challenges for training and evaluation (see Fig.~\ref{fig: introduction} (b)).
{\tt Challenge-Train}:
When all regions share the same brain age label, the model collapses to an over-averaged solution, obscuring regional differences and spatial gradients and thus degrading predictive performance.
{\tt Challenge-Eval}:
Without region-level ground truth, region-wise supervised evaluation is impossible, and standard metrics, such as mean absolute error (MAE), are inapplicable for assessing ReBA estimates.


To address {\tt Challenge-Train}, we propose the Regional Brain Age Prediction Network (\textbf{ReBA-Pred-Net}, see Fig.~\ref{fig: introduction} (c)), which estimates ReBA under whole-brain level weak supervision using a Teacher–Student framework.

\begin{figure*}[t]
\centering
\includegraphics[width=1.0\textwidth]{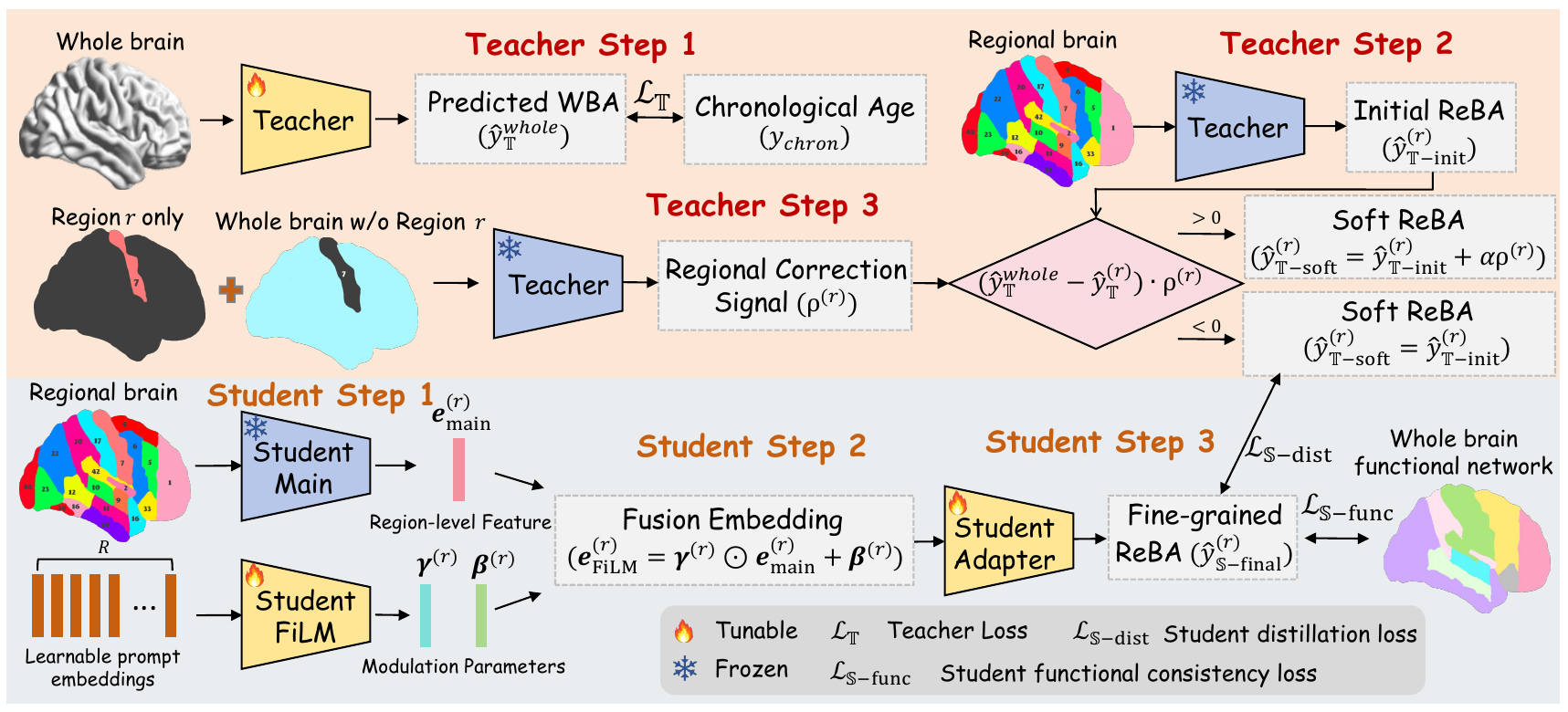} 
\vspace{-5mm}
\caption{ \small
The flowchart of Regional Brain Age Prediction Network (\textbf{ReBA-Pred-Net}).
\textbf{Teacher module} consists of three decoupled steps:
(i) Feed the whole-brain MRI to the Teacher to predict WBA and train it against chronological age;
(ii) Parcellate the brain using an atlas; extract each region and pass it through the frozen Teacher to obtain initial ReBA;
(iii) Apply the corrections in Eqs. (\ref{eqa: regional correction signal}), (\ref{eqa: teacher_final_regional_age}) to each region to produce the final soft ReBA.
\textbf{Student module} comprises three sequential steps:
(i) Feed each region to the Student Main (shared with the Teacher) to obtain a regional embedding;
(ii) Add a learnable prompt per region, generate modulated parameters with a tunable Student FiLM, and fuse them with the regional embedding;
(iii) Pass the fused feature through a lightweight adapter to produce the final fine-grained ReBA.
We optimize the Student with a distillation loss (to match the Teacher’s soft ReBA) and a functional-consistency loss (enforcing that regions within the same functional network change similarly).
\textbf{Notably, all training is conducted exclusively on HC.}
}
\label{fig: flowchart}
\end{figure*}

\textbf{(i)} The Teacher module (see Fig.~\ref{fig: flowchart} Top) converts weakly-supervised WBA into region-wise soft targets (\textit{directly optimizing based on the WBA invites shortcut learning and degenerate solutions}). 
After standard preprocessing (skull stripping, N4 bias correction, and nonlinear registration by DeepPrep~\cite{ren2025deepprep}), we train the Teacher to regress WBA using HC's chronological age and then freeze its weights. 
We then parcellate the brain with the atlas (\textit{e.g.}, Harvard–Oxford~\cite{jenkinson2012fsl}) and use the frozen Teacher on each region to obtain initial ReBA. 
Next, we apply an additive correction to each region according to its marginal contribution to the whole-brain prediction (\textit{i.e.}, the perturb-and-observe change in WBA after lightly occluding that region), yielding more reliable soft ReBA. This procedure mitigates over-averaging and turns global weak supervision into localized, teachable signals.

\textbf{(ii)} The Student module (see Fig.~\ref{fig: flowchart} Bottom) implements fine-grained regional estimation. On a shared backbone, we introduce a learnable prompt to each brain region, use it to perform Feature-wise Linear Modulation (FiLM), and employ a lightweight adapter to produce region-specific readouts. The Student is trained to match the Teacher’s corrected soft ReBA via a distillation loss. In addition, we impose a functional consistency constraint, encouraging regions within the same functional system (\textit{e.g.}, vision-related regions) to have similar ages, which suppresses over-averaging and preserves plausible spatial gradients.


For {\tt Challenge-Eval}, we present two complementary indirect metrics (see Fig.~\ref{fig: introduction} (c)): Healthy Control Similarity (\textbf{HCS}), assessing statistical consistency; and Neuro Disease Correlation (\textbf{NDC}), assessing factual consistency.

\textbf{(i)} HCS is evaluated on unseen HC. We compute the Regional Brain Age Gap ($\Delta$ReBA, ReBA minus chronological age) and compare its distribution against the HC training data. If the two distributions align, the model is well-calibrated on HC with no evident distributional drift. 
\textit{Note that, although this metric has limitations at the individual level, it captures group-level statistical consistency and thus serves as an effective gatekeeper for basic model reliability.}

\textbf{(ii)} NDC is evaluated on clinically confirmed neuro disorders. Using clinical priors to specify the disease-related region set (\textit{e.g.}, motor-related regions for PD, memory-related regions for AD, see Sec.~\ref{sec: Clinical Prior Knowledge}), we compute the $\Delta$ReBA individually within this set. 
When this gap is larger and significantly exceeds that of HC and non-target disease groups, the model’s ReBA pattern accords with the known disease prior, providing evidence for validity. 
\textit{Intuitively, in the absence of regional ground truth, we substitute “known facts” for “unknown labels”: NDC tests whether regions that should appear “older” indeed do so; satisfying this criterion supports the credibility of the model’s predictions.}



Our main contributions are summarized as follows:
\begin{itemize}
    \vspace{-4mm}
    \item 
    We identify the central obstacle in ReBA research as the lack of fine-grained region-level labels, and we cast this into two methodological challenges, \textit{i.e.}, {\tt Challenge-Train} and {\tt Challenge-Eval}.
    \vspace{-3mm}
    \item 
    We present ReBA-Pred-Net for {\tt Challenge-Train}, a Teacher–Student framework in which the Teacher, trained under whole-brain weak supervision, produces soft ReBA, and the Student employs region-specific prompts with consistency constraints, enabling reliable ReBA estimation despite weak supervision.
    \vspace{-3mm}
    \item 
    We design HCS and NDC for {\tt Challenge-Eval}. HCS emphasizes statistical consistency, whereas NDC emphasizes factual consistency; together they provide complementary, testable validation when region-level ground truth is unavailable. Empirical results demonstrate the effectiveness of our method.

\end{itemize}












\section{Related Work}



\textbf{Whole brain age (WBA).}
Research on WBA is relatively mature. Early work used voxel-based morphometry (VBM) features \cite{ashburner2000voxel,good2001voxel,pennanen2005voxel}, for example counts or summaries of gray and white matter at the voxel level, to regress age. These pipelines were limited by hand-crafted descriptors, sensitivity to smoothing and partial-volume effects, and site or scanner variability, which constrained accuracy. Subsequent studies adopted classical machine-learning regressors such as Gaussian processes \cite{cole2015prediction}, hidden Markov models \cite{wang2011mri}, and random forests \cite{liem2017predicting}, typically trained on VBM or global features extracted from T1-weighted MRI. 
With larger cohorts and greater compute, deep learning became the standard approach, where 3D CNNs and Transformer-based architectures take MRI as input and directly regress WBA \cite{lee2022deep,cheng2021brain,armanious2021age,jonsson2019brain,kuchcinski2023mri,yu2024brain,seitz2024brainage}. 
Evaluation protocols are well established, commonly reporting MAE, MSE, and Spearman’s rank correlation coefficient (SRCC) etc.


\textbf{Regional brain age (ReBA).}
\textit{In contrast to the mature field of WBA prediction, ReBA estimation remains at a nascent stage.}
Existing efforts to characterize regional aging fall predominantly into two categories: purely conceptual discussions~\cite{bethlehem2022brain,kalc2024brainage} and morphological feature-based methods~\cite{riccardi2025distinct,lee2022regional}. 
For the former, Bethlehem \textit{et al.} map normative trajectories of region-level morphometry across the human lifespan, and Kalc \textit{et al.} outline best practices for brain-age workflows. However, these studies focus on macro-level cohort statistics rather than providing an actionable predictive model for individual-level ReBA estimation.
For the latter, existing computational approaches rely on a two-stage paradigm using pre-extracted morphological features (\textit{e.g.}, cortical thickness and volume derived from FreeSurfer~\cite{fischl2012freesurfer}). This approach inherently suffers from significant spatial information loss (discarding texture and intensity gradients) and cumulative errors originating from segmentation and registration steps. Consequently, these models exhibit poor cross-site generalizability due to their high sensitivity to scanner variations and preprocessing pipelines.
These limitation underscores the urgent need for a general-purpose ReBA framework with standardized validation; our study addresses this by proposing ReBA-Pred-Net and introducing complementary metrics (HCS and NDC).



\section{Methodology}
\label{sec: Methodology}

\subsection{Overview}
\label{sec: overview}





We propose the Regional Brain Age Prediction Network (\textbf{ReBA-Pred-Net}), which comprises three components:

\textbf{(i)} Preprocessing module (Please see Sec.~\ref{sec: MRI Pre-processing}) applies DeepPrep~\cite{ren2025deepprep} for skull stripping, N4 bias correction and nonlinear registration on T1-weighted MRI to standardize anatomy \cite{avants2009advanced}.

\textbf{(ii)} Teacher module (Please see Fig.~\ref{fig: flowchart} Top) trains a WBA regressor with chronological age, freezes it, derives per-region initial ReBA with atlas parcellation, and then applies additive corrections to obtain soft ReBA.

\textbf{(iii)} Student module (Please see Fig.~\ref{fig: flowchart} Bottom) shares the backbone with the Teacher, extracts regional embeddings, injects learnable prompts and feature-wise linear modulation, fuses features, and uses a lightweight adapter to output fine-grained ReBA; it is optimized with distillation to soft ReBA plus a functional-consistency loss (regions in the same functional network should change similarly).

\subsection{MRI Pre-processing Module}

MRI scans vary in resolution, contrast, and signal-to-noise ratio across scanners, and also differ in anatomy between subjects. We therefore standardize all raw T1-weighted images with MRI-Processor to improve robustness and cross-subject comparability.
Define the raw MRI image as $\mathbf{X}_{\text{raw}} \in \mathbb{R}^{D \times H \times W}$, the workflow is:
\begin{align}
& \mathbf{X}_{\text{proc}}  =
        \mathcal{F}_{\text{DeepPrep}}
        \left(\mathbf{X}_{\text{raw}} 
        \right),
\label{eqa: pre_processing}
\end{align}
where $\mathbf{X}_{\text{proc}} \in \mathbb{R}^{D \times H \times W}$ denotes the processed MRI data, $D,H,W$ represent the depth, height, and width; 
$\mathcal{F}_{\text{DeepPrep}}$ denotes the operation of using DeepPrep to implement brain extraction, bias field correction, and nonlinear registration.

\subsection{Teacher Module}

\paragraph{Workflow.} 
Define the teacher module as $\pmb{\mathbb{T}}$.
Its purpose is to produce soft ReBA for fine-grained supervision of the Student via a five-step pipeline.
By converting coarse whole-brain supervision into regional signals, it avoids collapsing all regions to a single chronological-age label and enhances region-specific discrimination.
\textit{The key challenge is obtaining reliable soft labels, otherwise it is “garbage in, garbage out”.}
We first derive initial ReBA (Step iii), then refine it with an additive correction signal (Step iv), yielding the final soft ReBA used for supervision (Step v).
\textbf{Notably, all training is conducted exclusively on HC}.


\textbf{(i)} 
We begin by training the Teacher (model-agnostic) on the processed training data $\mathbf{X}_{\text{proc}}$ to predict WBA. It is trained by minimizing the discrepancy between this prediction and the chronological age. Upon convergence, the model’s parameters are frozen for subsequent stages. The estimated WBA can be formulated as:
\begin{align}
& {\hat{y}_{\pmb{\mathbb{T}}}^{(whole)}} =
        \pmb{\mathbb{T}}
        \left(\mathbf{X}_{\text{proc}} 
        \right),
\label{eqa: estimated whole-brain age}
\end{align}
where $\hat{y}_{\pmb{\mathbb{T}}}^{(whole)}$ denotes the estimated whole-brain age.

\textbf{(ii)} 
Define the atlas as $\mathbf{M}_{\text{raw}} \in \mathbb{R}^{D \times H \times W}$, where each voxel is assigned an integer label from $0$ to $R$, corresponding to one of $R$ brain regions, \textit{i.e.,} $\mathbf{M}_{\text{raw}}[i,j,k] \in \{0,1,2,\cdots,R\}$.
We perform one-hot encoding for the atlas, which can be formulated as:
\begin{align}
& \mathbf{M}_{\text{proc}} =
        \mathcal{F}_{\text{onehot}}
        \left(\mathbf{M}_{\text{raw}} 
        \right),
\label{eqa: onehot_atlas}
\end{align}
where 
$\mathbf{M}_{\text{proc}} \in \mathbb{R}^{R \times D \times H  \times W}$ denotes the one-hot atlas.

\textbf{(iii)}
Subsequently, the one-hot atlas is used to isolate voxel sets for each brain region. To reduce boundary artifacts, the brain region mask is dilated by $1$ voxel, and all non-selected regions are replaced with small noise ($\eta$). The process can be formulated as:
\begin{align}
& \mathbf{X}_{\text{proc}}^{(r)} =
        \mathbf{X}_{\text{proc}} \odot \mathbf{M}_{\text{proc}}^{(r)} + \eta \mathbf{Z}\odot\left(1-\mathbf{M}_{\text{proc}}^{(r)}\right),
\label{eqa: regional_input}
\end{align}
where $\mathbf{X}_{\text{proc}}^{(r)}$, $\mathbf{M}_{\text{proc}}^{(r)}$ denote the processed brain image and the atlas mask of the $r$-th region, respectively;
$\mathbf Z \sim \mathcal N(0,1)$, and $\eta \ll 1$.
Following, we use the Teacher to predict the initial ReBA of the $r$-th brain region by:
\begin{align}
& \hat{y}_{\pmb{\mathbb{T}}\text{-init}}^{(r)} = \pmb{\mathbb{T}}\left(\mathbf{X}_{\text{proc}}^{(r)}\right).
\label{eqa: initial_teacher_regional_age}
\end{align}

\textbf{(iv)} 
Meanwhile, we introduce a regional correction signal to quantify each brain region’s influence on WBA estimation. \textit{Intuitively, if the $r$-th region is “occluded” (e.g., replaced with small noise so that the teacher cannot perceive it), we measure how much the teacher’s WBA prediction changes. A positive value indicates that the region’s age should be higher than the WBA estimate; a negative value indicates it should be lower. This change serves as an auxiliary correction signal for ReBA prediction.}
Formally, let the occluded input be:
\begin{align}
& \mathbf{X}_{\text{proc}}^{(-r)} = 
\mathbf{X}_{\text{proc}} \odot \left(1-\mathbf{M}_{\text{proc}}^{(r)}\right) + \eta \mathbf{Z}\odot\mathbf{M}_{\text{proc}}^{(r)}.
\label{eqa: occluded input}
\end{align}
This regional correction signal can be formulated as:
\begin{align}
& \rho^{(r)} =
        \mathbb{E}_{\left(
        \mathbf{X}_{\text{proc}} \sim \mathcal{D}_\text{tr}
        \right)}
        \left[
        \pmb{\mathbb{T}} \left(\mathbf{X}_{\text{proc}} \right) - 
        \pmb{\mathbb{T}}  \left(\mathbf{X}_{\text{proc}}^{(-r)}  \right)
        \right], 
\label{eqa: regional correction signal}
\end{align}
where $\rho^{(r)} >0$ means the $r$-th brain region increases the predicted age, whereas
$\rho^{(r)} < 0$ lowers the prediction;
$\mathcal{D}_\text{tr}$ denotes the training data;
$\mathbb{E}_{(\cdot)}$ is the expectation operator;
$\mathbb{E}_{\left(\mathbf{X}_{\text{proc}} \sim \mathcal{D}_\text{tr}\right)}\left[ \cdot\right]$ indicates that, for every training example, we compute the correction signal for region $r$ and subsequently take the average across the dataset.


\textbf{(v)}
Finally, we correct the ReBA using this regional correction signal, which can be formulated as:
\begin{align}
& 
\hat{y}_{\pmb{\mathbb{T}}\text{-soft}}^{(r)} 
= \hat{y}_{\pmb{\mathbb{T}}\text{-init}}^{(r)} +
\mathds{1}_{\left((\hat{y}_{\pmb{\mathbb{T}}}^{(whole)}-\hat{y}_{\pmb{\mathbb{T}}\text{-init}}^{(r)}) \cdot \rho^{(r)} > 0 \right)} \cdot \alpha \rho^{(r)},
\label{eqa: teacher_final_regional_age}
\end{align}
where
$\mathds{1}_{(\text{condition})}$ is the indicator function, equal to $1$ if the condition is true and $0$ otherwise;
$\alpha$ is the hyperparameter.
\textit{Put simply, the $\mathds{1}_{\left((\hat{y}_{\pmb{\mathbb{T}}}^{(whole)}-\hat{y}_{\pmb{\mathbb{T}}\text{-init}}^{(r)}) \cdot \rho^{(r)} > 0 \right)}$ is used to check whether the direction of the predicted ReBA is consistent with the correction signal; if consistent, no update is applied, and otherwise, the prediction is adjusted toward the signal with a step size of $\alpha \rho^{(r)}$.}

\paragraph{Teacher Loss.}
During the teacher stage, we only minimize the MAE loss between the chronological age and the WBA predicted by the model, which can be formulated as:
\begin{align}
\mathcal{L}_{\pmb{\mathbb{T}}} = 
\frac{1}{N} \sum_{n=1}^N \left|\hat{y}_{\pmb{\mathbb{T}}}^{(whole,n)} - y_{\text{chron}}^{(n)}  \right|,
\label{eqa: loss_teacher}
\end{align}
where $\hat{y}_{\pmb{\mathbb{T}}}^{(whole,n)}, y_{\text{chron}}^{(n)}$ denote the predicted WBA and chronological age of the $n$-th subject.
After training, the teacher’s parameters are fully frozen. In the Eqs.~(\ref{eqa: initial_teacher_regional_age}) and (\ref{eqa: regional correction signal}), we directly invoke this frozen teacher to obtain the ReBA.

\subsection{Student Module}
\paragraph{Workflow.} 
Define the student module as $\pmb{\mathbb{S}}$, which performs on-demand ReBA prediction. It consists of three components: a teacher-inspired main block, 
a Feature-wise Linear Modulation (FiLM) block, and an adapter block, which can be formulated as: 
\begin{align}
\pmb{\mathbb{S}} = 
\left\{ 
\pmb{\mathbb{S}}_{\text{main}},\ \pmb{\mathbb{S}}_{\text{FiLM}}, \ \pmb{\mathbb{S}}_{\text{adapter}}
\right\}.
\label{eqa: student_module}
\end{align}
$\pmb{\mathbb{S}}_{\text{main}}$ shares the same backbone parameters as the teacher $\pmb{\mathbb{T}}$ but removes the regression head and remains frozen; 
$\pmb{\mathbb{S}}_{\text{FiLM}}$ and $\pmb{\mathbb{S}}_{\text{adapter}}$ are trainable.
The workflow has three steps:

\textbf{(i)}
First, following Eq.~(\ref{eqa: regional_input}), we extract the voxels corresponding to each brain region.
The resulting single region is then fed into the student main block to obtain a regional feature embedding. This yields a compact and stable region-level feature vector used for subsequent adapter-based prediction. We formulate the process as:
\begin{align}
& \mathbf{e}_{\text{main}}^{(r)} = 
\pmb{\mathbb{S}}_{\text{main}}\left(\mathbf{X}_{\text{proc}}^{(r)}\right),
\label{eqa: student_main_embedding}
\end{align}
where $\mathbf{e}_{\text{main}}^{(r)} \in \mathbb{R}^{d_{m}}$ is the $r$-th brain region embedding.


\textbf{(ii)}
Subsequently, for each brain region, we introduce a learnable prompt embedding ($\mathbf{p}^{(r)} \in \mathbb{R}^{d_p}$) and feed it into a lightweight FiLM block to produce the modulated parameters.
These parameters are then used to apply feature-wise affine modulation to the previously extracted regional feature embeddings.
\textit{Intuitively, the block learns how to generate region-specific readout rules from the regional prompt, thereby scaling and shifting each region’s features while sharing a single regression head. This design preserves region specificity, enables parameter sharing for efficiency, and fits naturally with a prompt-as-inference interface, yielding more stable and interpretable ReBA estimates.}
The process can be formulated as:
\begin{align}
& \gamma^{(r)},\beta^{(r)} =  
\pmb{\mathbb{S}}_{\text{FiLM}}\left(\mathbf{p}^{(r)}\right), \\
& \mathbf{e}_{\text{FiLM}}^{(r)} \ = 
\gamma^{(r)} \odot \mathbf{e}_{\text{main}}^{(r)} + \beta^{(r)},
\label{eqa: student_fusion_embedding}
\end{align}
where $\gamma^{(r)},\beta^{(r)} \in \mathbb{R}^{d_m}$ are the modulation parameters;
$\mathbf{e}_{\text{FiLM}}^{(r)} \in \mathbb{R}^{d_m}$ indicates the fusion embedding;
$\pmb{\mathbb{S}}_{\text{FiLM}}$ is implemented as a multilayer perceptron (MLP).

\textbf{(iii)}
Finally, based on the fused regional features, we build a single lightweight adapter shared across all brain regions to output the final ReBA:
\begin{align}
\hat{y}_{\pmb{\mathbb{S}}\text{-final}}^{(r)}= 
\pmb{\mathbb{S}}_{\text{adapter}} \left( \mathbf{e}_{\text{FiLM}}^{(r)} \right),
\label{eqa: student_final_regional_age}
\end{align}
where $\pmb{\mathbb{S}}_{\text{adapter}}$ is realized as a MLP.

\vspace{-5mm}
\paragraph{Student Loss}

The student loss comprises two components, which can be formulated as:
\begin{align}
\mathcal{L}_{\pmb{\mathbb{S}}} = 
\mathcal{L}_{\pmb{\mathbb{S}}\text{-dist}} + \zeta\ \mathcal{L}_{\pmb{\mathbb{S}}\text{-func}},
\label{eqa: loss_student_all}
\end{align}
where $\zeta$ is the hyperparameter.

\textbf{(i)} $\mathcal{L}_{\pmb{\mathbb{S}}\text{-dist}}$ indicates the distillation loss, encouraging the student’s ReBA predictions to fit the teacher’s soft ReBA, which can be formulated as:
\begin{align}
\mathcal{L}_{\pmb{\mathbb{S}}\text{-dist}} = 
\frac{1}{R \times N} \sum_{r=1}^R \sum_{n=1}^N 
\left|\hat{y}_{\pmb{\mathbb{S}}\text{-final}}^{(r,n)} - \hat{y}_{\pmb{\mathbb{T}}\text{-soft}}^{(r,n)}  \right|,
\label{eqa: loss_student_distiall}
\end{align}
where $\hat{y}_{\pmb{\mathbb{S}}\text{-final}}^{(r,n)}$, $\hat{y}_{\pmb{\mathbb{T}}\text{-soft}}^{(r,n)}$ denote the predicted ages for subject $n$ and region $r$, obtained from the student module and the teacher module, respectively.

\textbf{(ii)} $\mathcal{L}_{\pmb{\mathbb{S}}\text{-func}}$ indicates the functional consistency loss.
\textit{Motivated by clinical priors, brain regions belonging to the same functional network typically exhibit similar aging rates; large discrepancies within a network are uncommon (see Sec.~\ref{sec: Clinical Prior Knowledge}).
Based on this prior, for each subject $n$, we first compute the network-wise mean regional age and then penalize deviations of regions within the same network from this mean.}
The loss can be formulated as:
\begin{align}
& \hat{\mu}^{(k,n)} = \frac{1}{|\mathcal{G}^{(k)}|} \sum_{r \in \mathcal{G}^{(k)}} \hat{y}_{\pmb{\mathbb{S}}\text{-final}}^{(r,n)},\\
&\mathcal{L}_{\pmb{\mathbb{S}}\text{-func}} = 
\frac{1}{N} \sum_{n=1}^{N} \sum_{k=1}^{K}
 \frac{1}{|\mathcal{G}^{(k)}|} \sum_{r \in \mathcal{G}^{(k)}}
\left|\hat{y}_{\pmb{\mathbb{S}}\text{-final}}^{(r,n)} - \hat{\mu}^{(k,n)}  \right|,
\label{eqa: loss_student_func}
\end{align}
where 
$\mathcal{G}^{(k)}$ denotes the $k$-th brain network, 
and $\hat{\mu}^{(k,n)}$ denotes, for subject $n$, the mean predicted age across the regions contained in network $k$.



\section{Metric}
\label{sec: Metric}


As outlined in {\tt Challenge-Eval}, the lack of region-level ground truth makes conventional evaluation infeasible. We therefore introduce Healthy Control Similarity (HCS) to assess statistical consistency and Neuro-Disease Correlation (NDC) to assess factual consistency, providing an indirect yet testable validation. \textit{Notably, the two metrics are complementary: HCS measures population-level consistency on unseen HCs but is less sensitive to region-wise mode collapsing, whereas NDC probes disease-implicated regions (e.g., PD, AD), confirming the expected elevations versus HCs and exposing regional gradients that HCS may miss.}



\subsection{Healthy Control Similarity}
\label{sec: HCS}


The training data consists solely of HC, denoted $\mathcal{D}_\text{tr}^{\text{HC}}$.
At test stage, we use the unseen HC during training, denoted $\mathcal{D}_\text{ts}^{\text{HC}}$ ($\mathcal{D}_{\text{tr}} ^{\text{HC}}\cap \mathcal{D}_{\text{ts}}^{\text{HC}}= \emptyset$). 
For each brain region, we compute the $\Delta$ReBA on $\mathcal{D}_\text{ts}^{\text{HC}}$ and statistically compare its distribution with the corresponding distribution from $\mathcal{D}_\text{tr}^{\text{HC}}$.

Formally, we compute the regional brain age gap as:
\begin{align}
& \Delta\text{ReBA}^{(r,n)} = \hat{y}_{\pmb{\mathbb{S}}\text{-final}}^{(r,n)} - y_{\text{chron}}^{(n)},
\label{eqa: regional_brain_age_gap}
\end{align}
where $\Delta\text{ReBA}^{(r,n)}$ is the region $r$'s brain age gap for subject $n$. 
Let $\mathcal{H}_\text{tr}^{(r)} =\{ \Delta\text{ReBA}^{(r,n)}, n \in \mathcal{D}_\text{tr}^{\text{HC}}\}$ and $\mathcal{H}_\text{ts}^{(r)} =\{ \Delta\text{ReBA}^{(r,n)}, n \in \mathcal{D}_\text{ts}^{\text{HC}}\}$ denote the training and test distributions of brain age gaps for region $r$.
Define the per-region HCS via a statistical similarity function as:
\begin{align}
& \text{HCS}^{(r)} = 1 - \Phi_{\text{MMD}} \left(\mathcal{H}_\text{tr}^{(r)}, \mathcal{H}_\text{ts}^{(r)} \right),
\label{eqa: region-wise consistency score}
\end{align}
where $\Phi_{\text{MMD}}$ is the function of Maximum Mean Discrepancy (MMD) (see Sec.~\ref{sec: Bias Correction} for more details).
$\text{HCS}^{(r)} \in [0,1]$.
\textbf{Higher HCS indicates better distributional alignment and, consequently, more accurate predictions.}




\subsection{Neuro Disease Correlation}
\label{sec: NDA}




Guided by clinical knowledge and prior literature \cite{sarasso2021progression,liu2020brain}, we predefine disease-associated brain regions (\textit{e.g.}, regions 3,4,7,26 for PD, please see Sec.~\ref{sec: Clinical Prior Knowledge} Tab.~\ref{tab: disease_relevance} for more details). 
Patients with the relevant disorder typically show accelerated atrophy in the corresponding regions, causing the predicted ReBA of the disease-associated brain regions to exceed chronological age.
Accordingly, for patients with a confirmed diagnosis, NDC computes the $\Delta$ReBA between the corresponding regional brain age and the patient’s chronological age according to Eq.~(\ref{eqa: regional_brain_age_gap}).

Define the disease of patient as $\mathcal{D}_\text{ts}^{\text{Disease}} \in \{\mathcal{D}_\text{ts}^{\text{PD}}, \mathcal{D}_\text{ts}^{\text{AD}},\cdots\}$, the set of disease-associated brain regions is:
\begin{align}
& \mathcal{R}_\text{ts}^{\text{Disease}} = \text{Select}(\mathcal{D}_\text{ts}^{\text{Disease}}).
\label{eqa: select_associated_brain_regions}
\end{align}
The per-subject NDC can be formulated as: 
\begin{align}
& \text{NDC}^{(n)} = \frac{1}{|\mathcal{R}_\text{ts}^{\text{Disease}}|} 
\sum_{r \in \mathcal{R}_\text{ts}^{\text{Disease}}}
\frac{1}{1+e^{-\Delta \text{ReBA}^{(r,n)}}}.
\label{eqa: age_bias_correction}
\end{align}
$\text{NDC}^{(n)} \in [0,1]$.
\textbf{Higher NDC indicates stronger alignment with disease-characteristic regional patterns and, consequently, more credible ReBA estimates.}


\begin{figure*}[t]
\centering
\includegraphics[width=1.0\textwidth]{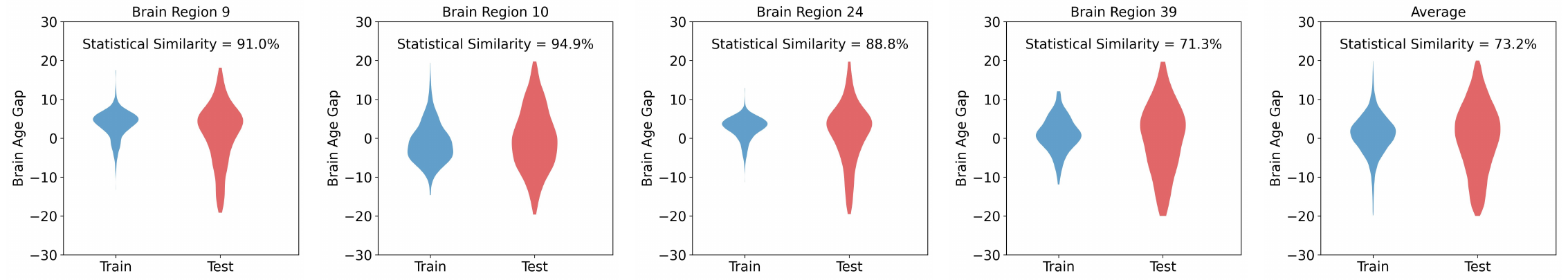} 
\vspace{-5mm}
\caption{ \small
Healthy Control Similarity (HCS) across brain regions (see Sec.~\ref{sec: Supplemented Experiments} for more regions' results). 
Results are based on the 3D DenseNet backbone \cite{lee2022deep}.
Blue bars denote the $\Delta$ReBA distributions predicted by our model on training HC; red bars denote the corresponding predictions on test HC. 
Statistical similarity is computed via Eq. (\ref{eqa: region-wise consistency score}). Higher is better. 
The results show consistently high statistical similarity at the regional level, with an overall HCS reaching 73\%, which supports the effectiveness of our method.
}
\label{fig: Experiments_HCS}
\end{figure*}

\begin{figure*}[t]
\centering
\includegraphics[width=1.0\textwidth]{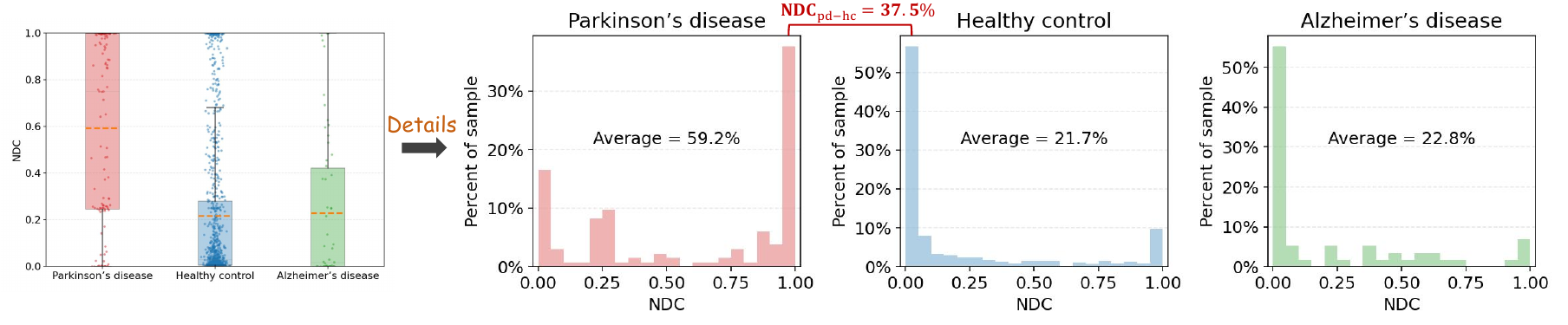} 
\vspace{-5mm}
\caption{ \small
Neuro-Disease Correlation (NDC) for Parkinson’s disease (PD). 
We focus on regions 3, 4, 7, 26, all of which show clinical evidence of accelerated aging in PD.
Accordingly, the $\Delta$ReBA in these regions is expected to exceed those observed in healthy controls (HC) and Alzheimer’s disease (AD). 
We compute the NDC and compare it across PD, HC, and AD. 
As anticipated, PD exhibits a markedly higher NDC than both HC and AD, providing indirect support for the accuracy of the proposed ReBA prediction.
}
\label{fig: Experiments_NDC}
\end{figure*}

\section{Experiments}





\textbf{Data.}
The data are split into training and test sets (please see Sec.~\ref{sec: Supplemented Experiments} Tab.~\ref{tab: dataset}). 
The training set contains 6,530 raw T1-weighted MRIs from 17 public datasets, comprising HC only, comprising HC only. 
The test set has three parts: 
1,057 unseen HC from other 9 public datasets and our in-house collection, used for the HCS metric; 
326 PD cases from PPMI~\cite{marek2011parkinson} and our in-house collection, used for NDC; 
and 107 AD cases from ADNI~\cite{mueller2005alzheimer}, also used for NDC.
All data will be public.

\vspace{-1mm}
\textbf{Implementation.}
Our model was trained for 60 epochs using the AdamW optimizer, configured with an initial learning rate of $1\times 10^{-4}$ and a weight decay of $1\times 10^{-5}$. The learning rate schedule followed a cosine annealing strategy. Several key hyperparameters were involved in training, including $\alpha = 1$, $\eta = 0.1$, and $\zeta = 1$ (see Sec.~\ref{sec: Supplemented Experiments} for sensitivity analysis ). 
All experiments were conducted on an NVIDIA RTX vGPU. The training batch size was set to 4.
The code will be publicly.

\vspace{-1mm}
\textbf{Comparison on HCS Metric.}
HCS quantifies, for each region, how closely the $\Delta$ReBA distributions match between training and test sets;
higher similarity indicates better prediction for that region (see Fig. \ref{fig: Experiments_HCS}).
Using the Harvard–Oxford atlas \cite{jenkinson2012fsl}, we parcellate the brain into 48 regions. 
Fig.~\ref{fig: Experiments_HCS} reports similarities for four exemplar regions (see Sec.~\ref{sec: Supplemented Experiments} for more details) and the average results based on 3D DenseNet backbone \cite{lee2022deep}. 
The $\Delta$ReBA similarity for the same region is consistently high, exceeding 73\%, validating the effectiveness of our approach.


\vspace{-5mm}
\paragraph{Comparison on NDC Metric.}

NDC incorporates clinical priors (see Sec.~\ref{sec: Clinical Prior Knowledge}) into evaluation: for a given disease, it prespecifies the set of related brain regions and tests whether those regions that “should age earlier” indeed do so. Fig.~\ref{fig: Experiments_NDC} takes PD as an example (see Sec.~\ref{sec: Supplemented Experiments} for other diseases); clinically, PD is known to show accelerated aging in regions 3, 4, 7, and 26. Accordingly, we compute the NDC score of these four regions and take the average, and we compare its distribution across the PD, HC, and AD cohorts. The NDC for PD is about 60\%, markedly higher than 20\% for HC and AD, thereby indirectly confirming the validity of the ReBA estimates.

\begin{table}[!t]
\centering
\small
\setlength{\tabcolsep}{1.2mm}{
\begin{tabular}{ccccccccccccc}
\toprule 
\multirow{2}{*}{} 
& \multicolumn{2}{c}{\textbf{Teacher}} 
& \multicolumn{2}{c}{\textbf{Student}} 
& \multirow{2}{*}{\textbf{HCS}} 
& \multirow{2}{*}{\textbf{NDC}$_{\text{{pd}-{hc}}}$}
\\
\cmidrule(lr){2-3} \cmidrule(lr){4-5}
& Initial ReBA & Soft ReBA & FiLM & $\mathcal{L}_{\pmb{\mathbb{S}}\text{-func}}$ \\
\midrule
\ding{172} & & & \checkmark & \checkmark &17.9   & 17.5 \\
\ding{173} & \checkmark & & \checkmark & \checkmark & 67.5 & 28.6  \\
\ding{174} & \checkmark & \checkmark & &  & 71.3 & 6.2  \\
\ding{175} & \checkmark & \checkmark &  & \checkmark & 72.2 &  14.3 \\
\ding{176} & \checkmark & \checkmark & \checkmark & & 61.2 & 24.8  \\
\ding{177} & \checkmark & \checkmark & \checkmark & \checkmark & \textbf{73.2} & \textbf{37.5} \\ 
\bottomrule
\end{tabular}
}
\caption{\small Ablation study (\%).
HCS and NDC are the proposed metrics.
NDC$_{\text{{pd}-{hc}}}$ is computed on PD-implicated regions as the difference between the NDC of the PD cohort and that of HC.
}
\label{tab: ablation_study}
\end{table}

\begin{table}[!t]
\centering
\small
\setlength{\tabcolsep}{0.5mm}{
\begin{tabular}{llccccccccccc}
\toprule 
{\textbf{Backbone}} 
& {\textbf{Param}}
& {\textbf{Time}} 
& {\textbf{HCS}} 
& {\textbf{NDC}$_{\text{{pd}-{hc}}}$}
\\
\midrule
M3T~\cite{jang2022m3t} 
& 3.9M & 568.8min &17.5  &21.0  \\
ViT3D~\cite{chen2023masked} 
& 1.5M & 86.4min  &20.6  &15.3  \\
Swin3D~\cite{chen2023masked} 
& 7.3M & 133.2min & 18.2  & 15.1 \\
SwinUNETR~\cite{tang2022self} 
& 52.6M & 421.2min &18.3  & 1.34  \\
AE-FLOW~\cite{zhao2023ae}
& 56.9M & 218.4min &48.7  &20.4  \\
S3D~\cite{wald2025revisiting}
& 16.8M & 202.8min &65.3  &28.6  \\
3D DenseNet~\cite{lee2022deep} 
& 76.5M & 94.8min &\textbf{73.2} & \textbf{37.5} \\
\bottomrule
\end{tabular}
}
\caption{\small Comparison (\%) of different backbones.
Most results are reported with the 3D DenseNet.
M3T, ViT3D, Swin3D, and SwinUNETR are based on the Transformer architecture, AE-FLOW, S3D and 3D DenseNet are based on the CNN architecture. 
}
\label{tab: different_backbones}
\end{table}


\begin{figure}[t]
\centering
\includegraphics[width=0.45\textwidth]{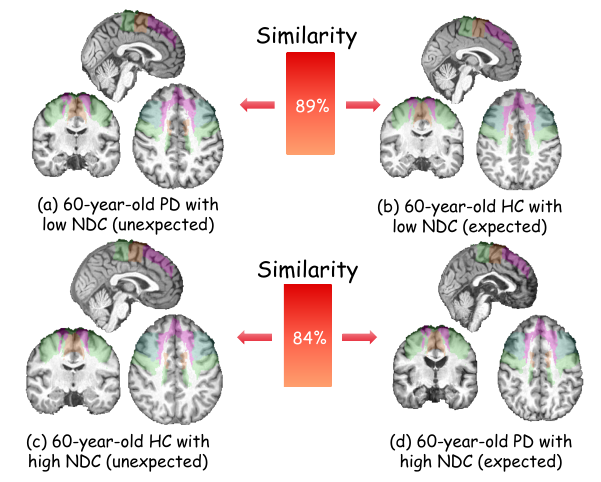} 
\vspace{-1mm}
\caption{ \small
Failure case visualization. 
(a) and (c) are failures: (a) PD with low NDC resembles the HC in (b); (c) HC with high NDC resembles the PD in (d). Morphometric similarity in PD-related regions likely drives the unexpected NDC.
}
\label{fig: failure_cases}
\end{figure}

\vspace{-5mm}
\paragraph{Ablation Studies.}





Our method comprises Teacher and Student components. We assess each module via ablation; detailed results appear in Tab.~\ref{tab: ablation_study}.
We report HCS and NDC$_{\text{{pd}-{hc}}}$, where is computed on PD-implicated regions as the difference between the NDC of the PD and HC.


For Teacher.
\textbf{(i)} Row \ding{172} supervises the Student directly with chronological age. Compared with Row \ding{177}, both HCS and NDC$_{\text{{pd}-{hc}}}$ drop sharply, indicating that coarse WBA labels cannot be used to predict ReBA and underscoring the necessity of the Teacher.
\textbf{(ii)} Row \ding{173} adopts the initial ReBA from Eq.~(\ref{eqa: initial_teacher_regional_age}) as the distillation target. Compared with Row \ding{177}, HCS and NDC$_{\text{{pd}-{hc}}}$ decline by about 7-9 \%, confirming that the additive regional correction is critical for producing high-quality soft targets for effective Student learning.

For Student.
\textbf{(i)} Row \ding{174} removes the Student entirely, leading to 2\% drop in HCS and a dramatic 30\% drop in NDC$_{\text{{pd}-{hc}}}$, underscoring the critical role of the Student.
\textbf{(ii)} Row \ding{175} excludes the FiLM block. While HCS remains similar, NDC$_{\text{{pd}-{hc}}}$ decreases by about 22 \%. FiLM introduces region-dependent affine modulation via learnable prompts, enabling individualized readout; without it, regional predictions become homogeneous, which has little effect on HCS but substantially harms NDC.
\textbf{(iii)} Row \ding{176} omits the functional-consistency loss. HCS and NDC decrease 12-13\%, demonstrating that this constraint helps suppress erratic variance while preserving meaningful gradients.


\vspace{-3mm}
\paragraph{Different Backbones.}
Our framework is model-agnostic and compatible with standard backbones. Most results are reported with the 3D DenseNet \cite{lee2022deep} (CNN).
We also evaluate alternative CNN backbones, including AE-FLOW \cite{zhao2023ae}, S3D \cite{wald2025revisiting}, as well as Transformer-based backbones, including M3T \cite{jang2022m3t}, ViT3D \cite{chen2023masked}, Swin3D \cite{chen2023masked}, SwinUNETR \cite{tang2022self}.
Comparison results are listed in Tab.~\ref{tab: different_backbones}.

\vspace{-3mm}
\paragraph{Failure Cases and Analysis.}





As shown in Fig.~\ref{fig: Experiments_NDC} (left 2), some PD patients exhibit unexpectedly low NDC values, while in the left 3 and 4, several HC and AD subjects show abnormally high NDC values. These observations deviate from our expectations, so we conducted a focused analysis.

Beyond the intrinsic limitations and uncertainty of the model itself, several factors may contribute to these atypical outcomes:
\textbf{(i)} Some participants have inaccurate or unreliable age records, directly affecting NDC.
\textbf{(ii)} Early-stage PD patients may show minimal structural change, yielding limited regional brain-age elevation and thus lower NDC. Fig.~\ref{fig: failure_cases} (a) depicts a 60-year-old PD patient with low NDC (unexpected), whereas Fig.~\ref{fig: failure_cases} (b) shows an age-matched HC with a similarly low NDC (expected). The two exhibit highly similar morphometric patterns in PD-related regions, which likely explains the PD case’s unexpected outcome.
\textbf{(iii)} Some HC or AD subjects may actually be prodromal or high-risk PD individuals, showing accelerated aging in PD-related regions and consequently elevated NDC values. Fig.~\ref{fig: failure_cases}(c) presents a 60-year-old HC with high NDC (unexpected), while Fig.~\ref{fig: failure_cases}(d) shows a 60-year-old PD patient with high NDC (expected). These two share highly similar morphometric patterns in PD-implicated regions, which explains the HC case’s unexpected outcome.

\vspace{-3mm}
\section{Discussion}


The aim of this work is to predict ReBA. In the absence of regional clinical ground truth, standard metrics (\textit{e.g.}, MAE) are inapplicable. Instead, we introduce two statistically sound and clinically aligned indirect metrics. This raises a core question: \textit{if the model performs well on these metrics, can it be used in practice?} \textbf{We believe yes.}
\textbf{The ultimate goal of ReBA is not to recover an absolute “true” regional age, but to reveal regional abnormalities, early risks, or developmental deviations.} Our model identifies condition-specific departures from expected aging patterns, helping clinicians flag regions that merit targeted follow-up and providing researchers with valuable scientific insights. Thus, even without direct ground-truth validation, the approach remains both scientifically and clinically impactful.



\vspace{-3mm}
\section{Conclusion and Future Work}
\paragraph{Conclusion.}
This paper presents the first complete framework for ReBA prediction. To address the lack of fine‑grained ground‑truth labels, we adopt a Teacher–Student training paradigm in which the Teacher generates soft labels to supervise the Student, yielding stable and interpretable ReBA estimates.
For validation, we introduce two complementary metrics (HCS and NDC) that constrain the model from the perspectives of statistical consistency and factual consistency. Experimental results demonstrate the effectiveness of the proposed method.
\vspace{-5mm}
\paragraph{Limitations and Future Work.}
\textbf{(i)} Training. We observe residual failure cases failure cases, many of which cannot yet be clearly ascribed to model limitations versus external confounders, hindering wider deployment. Future work will integrate clinical expertise, tighten data quality control, and incorporate uncertainty estimation and cross-site calibration to further improve accuracy.
\textbf{(ii)} Validation. Current validation is necessary but not sufficient: NDC depends on clinical priors, and robust priors and data are scarce beyond a few diseases (\textit{e.g.}, PD, AD). To broaden applicability without increasing labeling burden, we plan to develop more accessible validation routes by leveraging longitudinal follow-up and routinely collected clinical scales and imaging biomarkers to derive finer-grained, severity-aware indices as proxy ground truths, thereby keeping validation cost-conscious while enabling more clinically meaningful evaluation.

\appendix

\begin{center}
{\Large \bf Appendix}
\end{center}

\section{Abbreviation}

This paper contains numerous abbreviations. For ease of reading, important abbreviations are listed in Tab.~\ref{tab: Abbreviations}.

\begin{table*}[h]
\centering
\renewcommand{\arraystretch}{1.1}
\setlength{\tabcolsep}{4pt}
\begin{tabular}{c|l|l}
\hline
\textbf{No.} & \textbf{Abbreviation} & \textbf{Full Name} \\
\hline
1  & WBA                    & Whole brain age        \\
2  & ReBA                   & Regional brain age        \\
3  & $\Delta$ReBA           & Regional brain age gap, \textit{i.e.}, ReBA minus chronological age       \\
4  & HCS                    & Metric of Healthy Control Similarity        \\
5  & NDC                    & Metric of Neuro Disease Correlation  \\
6  & HC                     & Healthy Control  \\
7  & PD                     & Parkinson’ disease       \\
8  & AD                     & Alzheimer’s disease        \\
9  & MRI                    & Magnetic Resonance Imaging         \\
10 & MAE                    & Metric of Mean absolute error       \\
11 & MSE                    & Metric of Mean squared error \\
12 & SRCC                   & Metric of Spearman’s rank correlation coefficient \\
13 & FiLM                   & Feature-wise Linear Modulation        \\
\hline
\end{tabular}
\caption{\small Abbreviations.}
\label{tab: Abbreviations}
\end{table*}

\begin{figure*}[ht]
\centering
\includegraphics[width=1.0\textwidth]{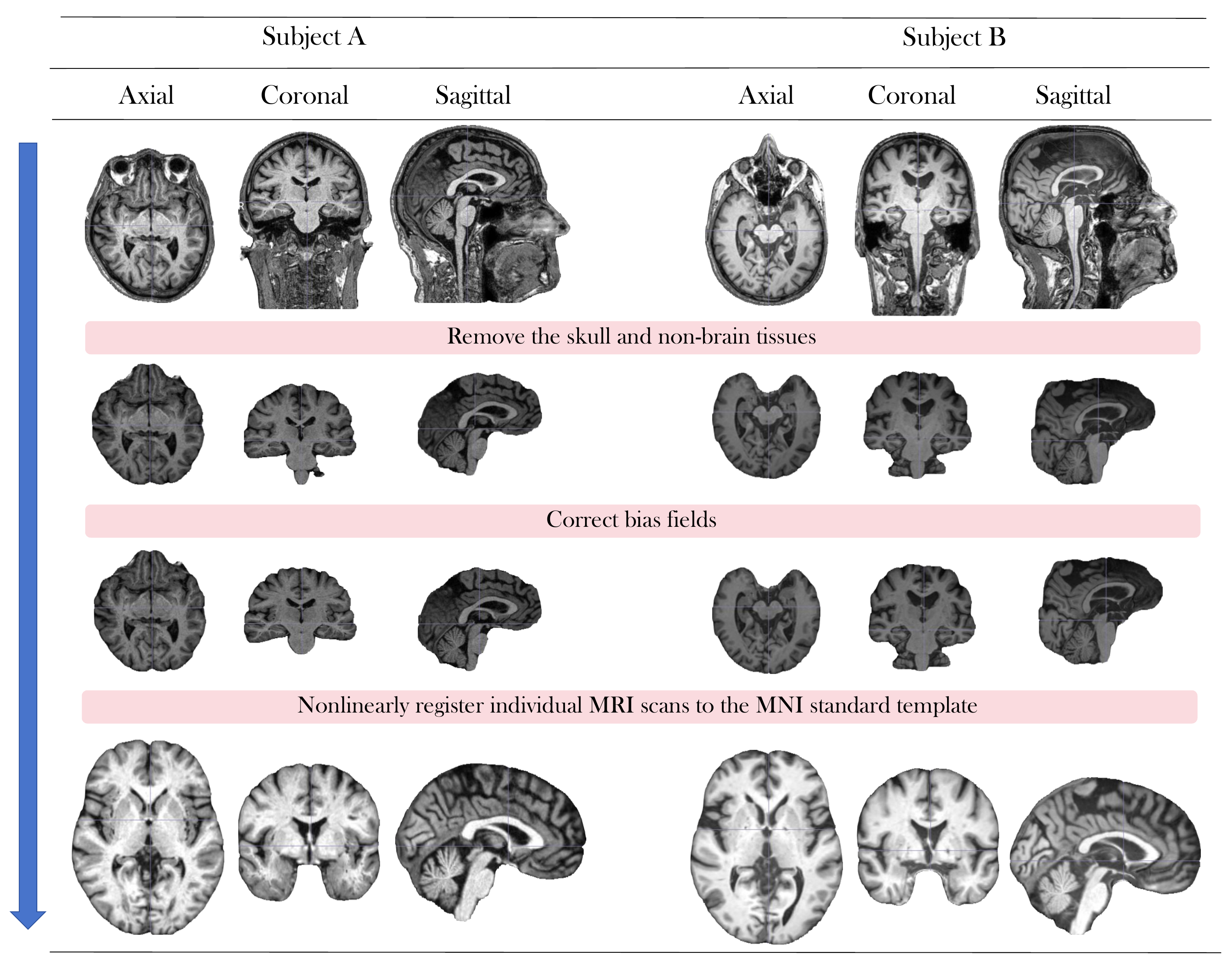} 
\caption{ \small
Flowchart of MRI Pre-processing.
}
\label{fig: Flowchart of MRI-Processor}
\end{figure*}

\begin{figure*}[ht]
\centering
\includegraphics[width=0.9\textwidth]{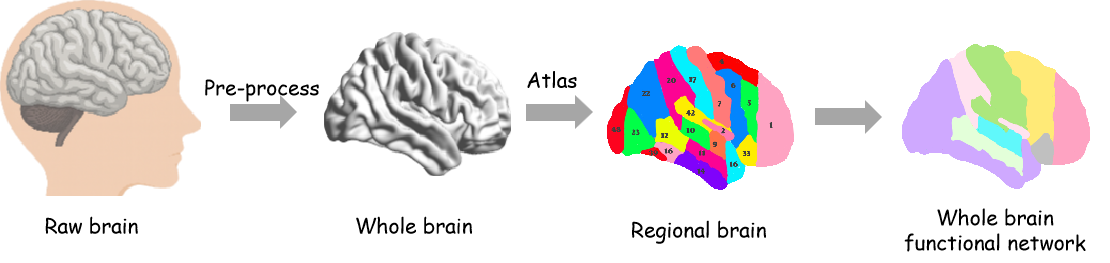} 
\caption{ \small
Illustration of the mapping from atlas-defined brain regions to functionally coherent brain networks.
}
\label{fig: brain_region_brain_network}
\end{figure*}

\section{Supplemented Related Work}

Brain age has emerged as a widely used imaging biomarker that captures the apparent biological aging of an individual’s brain relative to their chronological age. It provides a compact, interpretable indicator with broad utility in population health assessment, early disease screening, risk stratification, and longitudinal monitoring of neurodegeneration. As large neuroimaging cohorts and deep learning continue to evolve, estimating brain age from structural MRI has become a key task in computational neuroscience.

\textbf{Whole brain age (WBA).}
Research on WBA is relatively mature. Early work relied on voxel-based morphometry (VBM) features \cite{ashburner2000voxel,good2001voxel,pennanen2005voxel}, where voxelwise gray- and white-matter measures were summarized into handcrafted descriptors and fed into regression models. Despite their influence, VBM-based methods were constrained by sensitivity to registration quality, smoothing parameters, partial-volume effects, and scanner variability—issues that undermined predictive accuracy and cross-site robustness. Subsequent studies explored classical machine-learning regressors such as Gaussian processes \cite{cole2015prediction}, hidden Markov models \cite{wang2011mri}, and random forests \cite{liem2017predicting}, typically using VBM or global T1-MRI features.
With the advent of large datasets and modern compute, deep learning became the dominant paradigm. Recent approaches based on 3D CNNs and Transformer-style architectures directly ingest volumetric MRI and learn age-related patterns end-to-end \cite{lee2022deep,cheng2021brain,armanious2021age,jonsson2019brain,kuchcinski2023mri,yu2024brain,seitz2024brainage}. While WBA protocols are now standardized, the metric inherently collapses all spatial information into a single scalar. This global averaging dilutes regionally specific signals—crucial for heterogeneous conditions like Parkinson’s or Alzheimer’s disease—masking disease-relevant deviations and limiting interpretability.

\textbf{Regional brain age (ReBA).}
In contrast to the mature field of WBA, ReBA estimation remains at a nascent stage. Despite the clear potential of regional granularity, the existing literature, for examples,  \cite{kaufmann2019common,taylor2022investigating,lee2022regional,busby2024regional,bethlehem2022brain}, predominantly relies on a \textit{feature-based, two-stage paradigm}.
In these frameworks, regional descriptors (e.g., cortical thickness, subcortical volumes) must first be derived using morphometric packages like FreeSurfer before being fed into downstream statistical models or regressors. While these studies have successfully highlighted that disease-specific aging patterns exist and carry clinical value, their reliance on pre-extracted scalar features introduces a fundamental bottleneck.
As noted in the main text, this approach inherently discards rich, high-dimensional spatial information (such as local texture and intensity gradients) and binds the final performance to the success of complex preprocessing pipelines. Even recent extensions, such as the lobar-level predictions by Kalc \emph{et al.} \cite{bethlehem2022brain} or the post-hoc disparity indices by Wu \emph{et al.} \cite{wu2025regional}, operate within or atop these feature-constrained limitations.
Consequently, the field lacks a general-purpose framework capable of learning regional aging patterns directly from voxel-wise MRI data in an end-to-end manner, a gap our ReBA-Pred-Net is designed to fill.


\begin{figure*}[ht]
\centering
\includegraphics[width=1.0\textwidth]{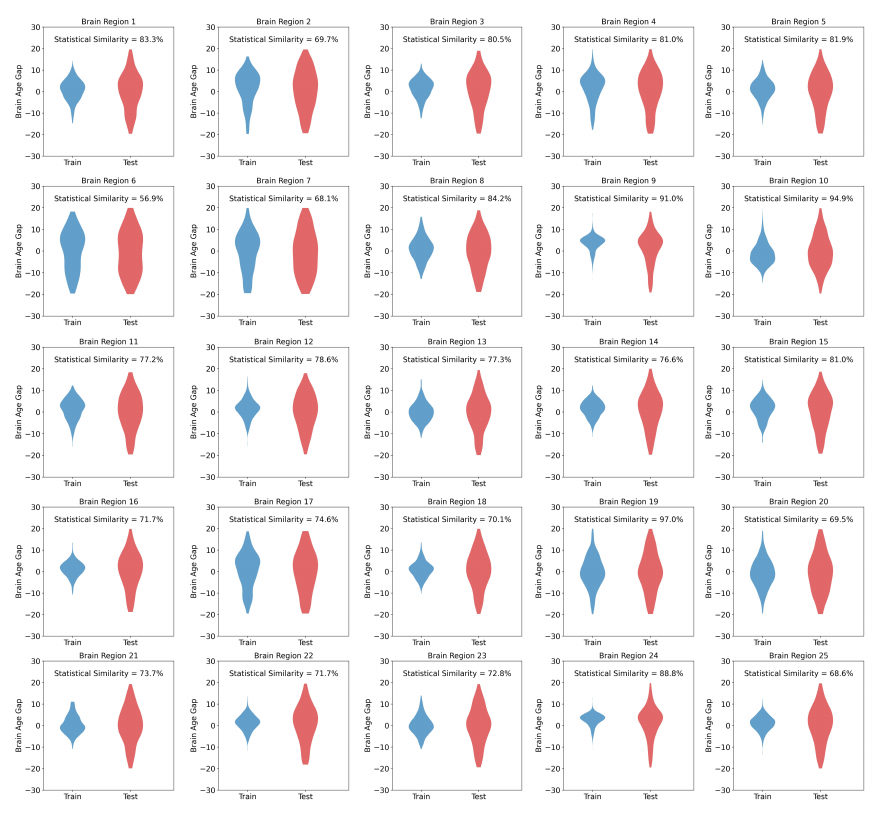} 
\vspace{-7mm}
\caption{ \small
Healthy Control Similarity (HCS) across brain regions (1-25 regions). 
Results are based on the 3D DenseNet backbone \cite{lee2022deep}.
Blue bars denote the $\Delta$ReBA distributions predicted by our model on training HC; red bars denote the corresponding predictions on test HC. 
Statistical similarity is computed via Main. Eq. (20). Higher is better. 
}
\label{fig: HCS_all_regions_1-25}
\end{figure*}

\begin{figure*}[ht]
\centering
\includegraphics[width=1.0\textwidth]{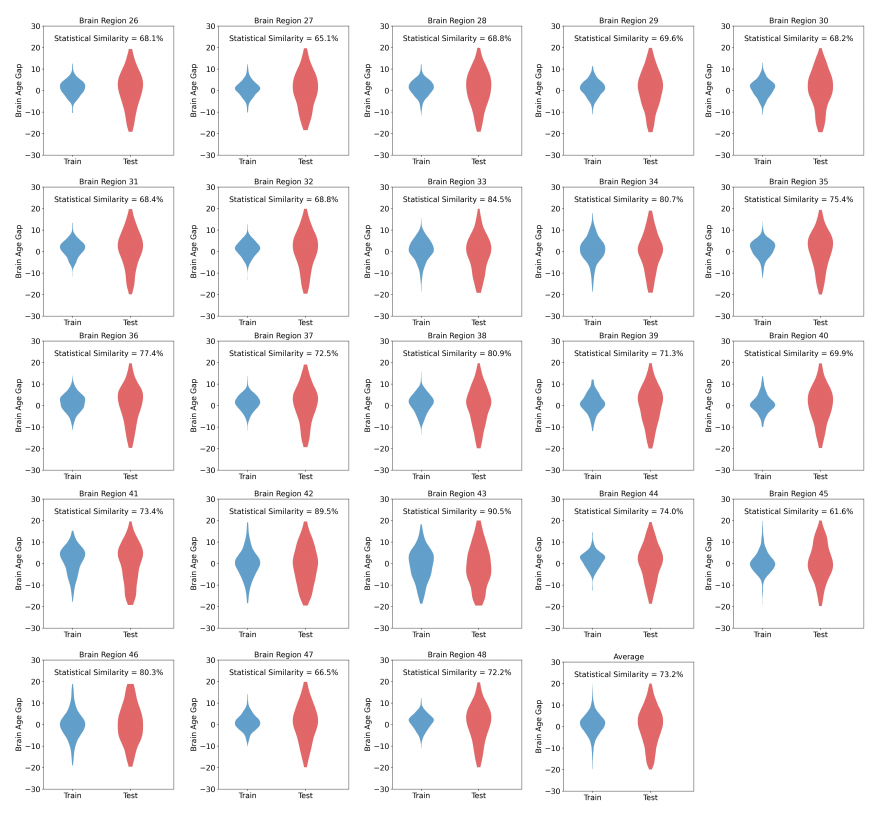} 
\vspace{-5mm}
\caption{ \small
Healthy Control Similarity (HCS) across brain regions (26-48 regions and average). 
Results are based on the 3D DenseNet backbone \cite{lee2022deep}.
Blue bars denote the $\Delta$ReBA distributions predicted by our model on training HC; red bars denote the corresponding predictions on test HC. 
Statistical similarity is computed via Main. Eq. (20). Higher is better. 
}
\label{fig: HCS_all_regions_26-48}
\end{figure*}

\begin{figure*}[ht]
\centering
\includegraphics[width=1.0\textwidth]{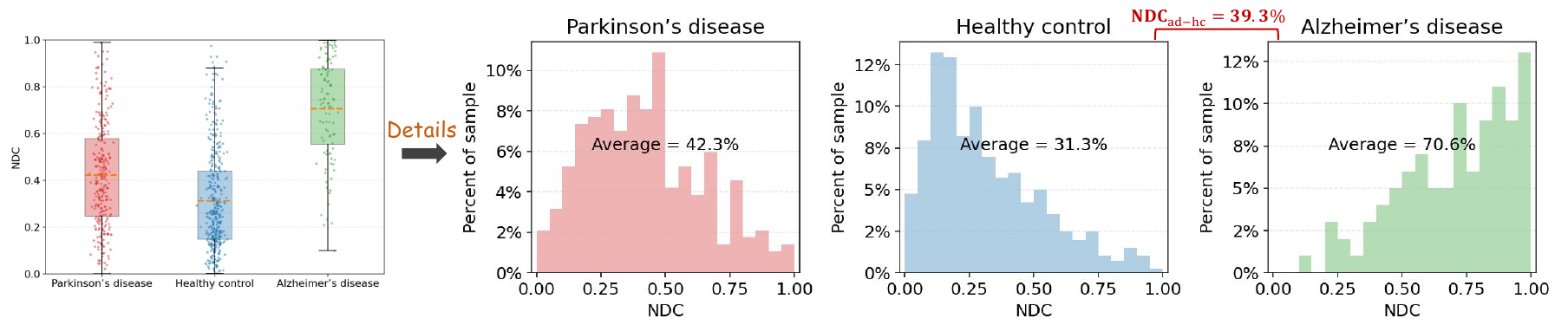} 
\vspace{-3mm}
\caption{ \small
Neuro-Disease Correlation (NDC) for Alzheimer’s disease (AD). 
We focus on regions 8-16,20-23,30,31,34,35,37-39, all of which show clinical evidence of accelerated aging in AD.
Accordingly, the $\Delta$ReBA in these regions is expected to exceed those observed in healthy controls (HC) and Parkinson’s disease (PD). 
We compute the NDC and compare it across AD, HC, and PD. 
As anticipated, AD exhibits a markedly higher NDC than both HC and PD, providing indirect support for the accuracy of the proposed ReBA prediction.
}
\label{fig: Experiments_AD_NDC}
\end{figure*}

\begin{figure*}[ht]
\centering
\includegraphics[width=1.0\textwidth]{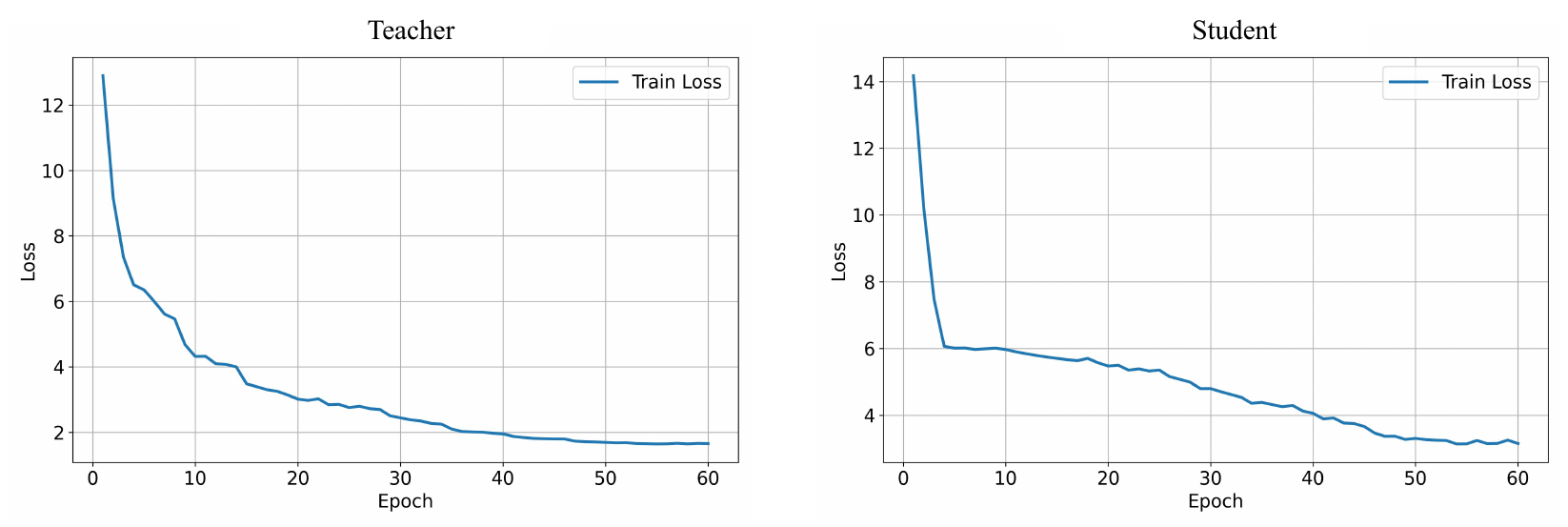} 
\vspace{-5mm}
\caption{ \small
Training loss of Teacher and Student modules.
}
\label{fig: training_curve}
\end{figure*}

\begin{figure*}[h]
\centering
\includegraphics[width=1.0\textwidth]{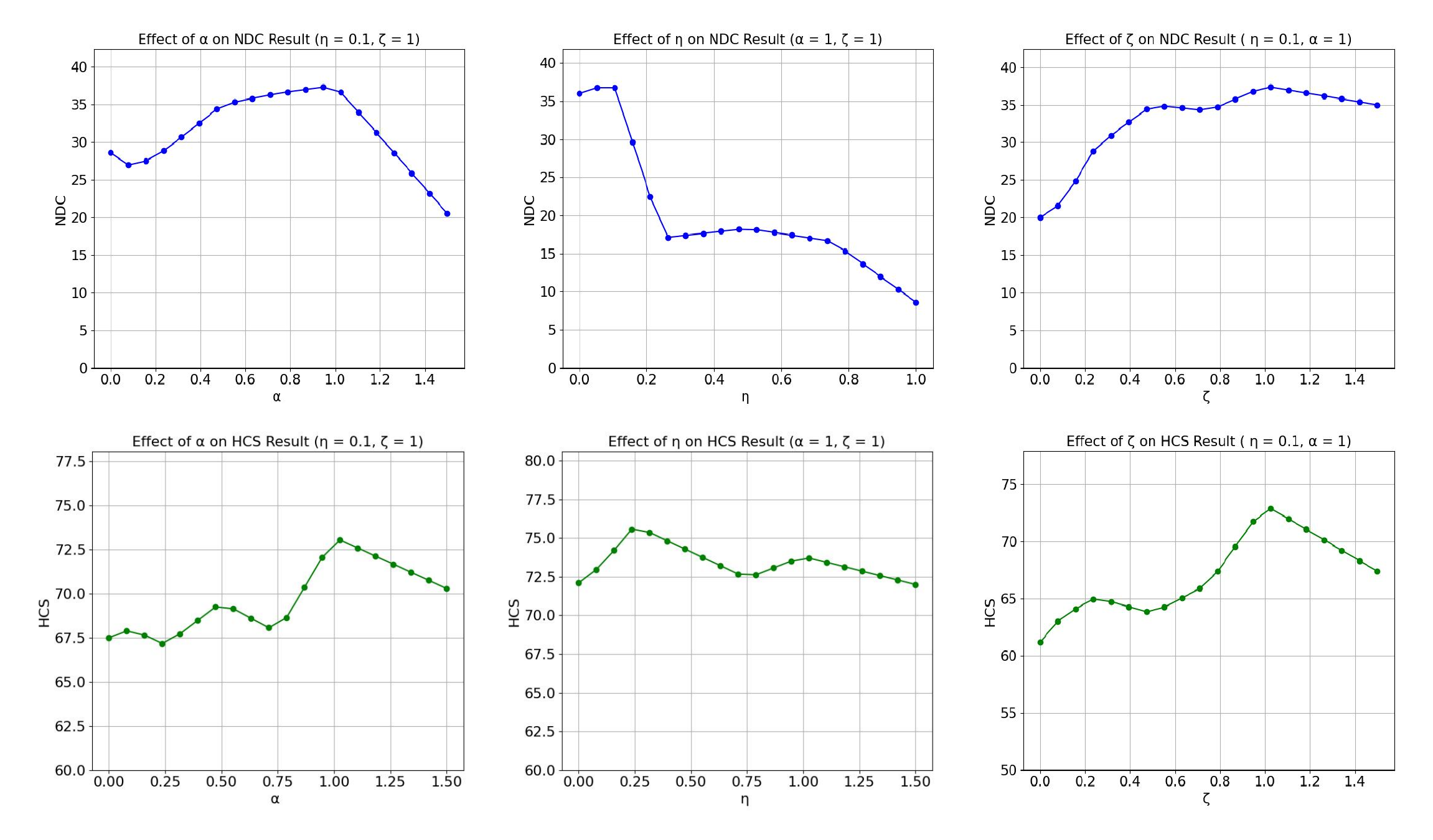} 
\vspace{-7mm}
\caption{ \small
Sensitivity of hyperparameters.
}
\label{fig: hyperparameters}
\end{figure*}

\section{MRI Pre-processing}
\label{sec: MRI Pre-processing}
Due to variations in resolution, contrast, and signal-to-noise ratio across MRI scanners, as well as substantial inter-subject differences in brain size, shape, and anatomical structure, we propose MRI-Processor to standardize all raw T1-weighted MRI scans through a three-step pre-processing pipeline to enhance the robustness of downstream modeling and improve cross-subject comparability:
(1) Brain extraction: Remove the skull and non-brain tissues to isolate the brain parenchyma, reducing irrelevant background noise;
(2) Bias field correction: Correct intensity inhomogeneities caused by magnetic field non-uniformities, improving overall image intensity uniformity and comparability; and
(3) Nonlinear registration: Align each MRI scan to the Montreal Neurological Institute (MNI) standard template to reduce anatomical variability between subjects and facilitate cross-subject analysis and model generalization.
This entire workflow was implemented using mature open-source toolkits.
We use the DeepPrep~\cite{ren2025deepprep} to complete these steps.
The flowchart is shown in Fig. \ref{fig: Flowchart of MRI-Processor}.

\section{Clinical Prior Knowledge}
\label{sec: Clinical Prior Knowledge}

The human brain can be subdivided into multiple regions according to anatomical and functional characteristics. In this study, we use the Harvard–Oxford Atlas \citep{jenkinson2012fsl} for standardized parcellation, ensuring consistent region definitions across all subjects. Regions that share similar functional roles can further be grouped into brain networks, which reflect coordinated activity and shared neurobiological functions. A schematic illustration of these region-to-network mappings is shown in Fig.~\ref{fig: brain_region_brain_network}.

Neurodegenerative diseases typically do not affect the brain uniformly. In Parkinson’s disease (PD), pathological changes are concentrated in regions involved in motor control, executive function, and affective regulation, while other regions remain relatively preserved. Alzheimer’s disease (AD), in contrast, exhibits early and prominent degeneration in memory-related and association cortices—such as the medial temporal lobe, posterior cingulate cortex, and parietal association areas—before spreading to broader cortical systems. This spatial heterogeneity has been consistently reported in prior neuropathological and neuroimaging studies \cite{gao2016study,burciu2018imaging}.

Building on clinical expertise and existing evidence, we categorize each atlas-defined region into one of three relevance levels for PD and AD: strongly-associated, potentially-associated, and non-associated. These assignments, summarized in Tab.~\ref{tab: disease_relevance}, serve as disease-informed priors that guide the interpretation of region-specific aging patterns in our analyses.

\begin{table*}[t]
\centering
\renewcommand{\arraystretch}{0.9}
\setlength{\tabcolsep}{3pt}
\begin{tabular}{c|l|c|c}
\hline
\textbf{Brain Region No.} & \textbf{Brain Region Name} & \textbf{Relevance with PD} & \textbf{Relevance with AD} \\
\hline
1  & Frontal Pole                                & None            & Potentially   \\
2  & Insular Cortex                              & Potentially     & None          \\
3  & Superior Frontal Gyrus                      & Strong          & Potentially   \\
4  & Middle Frontal Gyrus                        & Strong          & Potentially   \\
5  & Inferior Frontal Gyrus, Triangular Part     & Potentially     & Potentially   \\
6  & Inferior Frontal Gyrus, Opercular Part      & Potentially     & Potentially   \\
7  & Precentral Gyrus                            & Strong          & None          \\
8  & Temporal Pole                               & None            & Strong        \\
9  & Superior Temporal Gyrus, Anterior Division  & None            & Strong        \\
10 & Superior Temporal Gyrus, Posterior Division & None            & Strong        \\
11 & Middle Temporal Gyrus, Anterior Division    & None            & Strong        \\
12 & Middle Temporal Gyrus, Posterior Division   & None            & Strong        \\
13 & Temporooccipital Middle Temporal Gyrus      & None            & Strong        \\
14 & Inferior Temporal Gyrus, Anterior Division  & None            & Strong        \\
15 & Inferior Temporal Gyrus, Posterior Division & None            & Strong        \\
16 & Temporooccipital Inferior Temporal Gyrus    & None            & Strong        \\
17 & Postcentral Gyrus                           & Potentially     & None          \\
18 & Superior Parietal Lobule                    & Potentially     & None          \\
19 & Supramarginal Gyrus, Anterior Division      & None            & None          \\
20 & Supramarginal Gyrus, Posterior Division     & None            & Strong        \\
21 & Angular Gyrus                               & Potentially     & Strong        \\
22 & Lateral Occipital Cortex, Superior Division & None            & Strong        \\
23 & Lateral Occipital Cortex, Inferior Division & None            & Strong        \\
24 & Intracalcarine Cortex                       & None            & None          \\
25 & Medial Frontal Cortex                       & Potentially     & Potentially   \\
26 & Juxtapositional Lobule Cortex (SMA)         & Strong          & None          \\
27 & Subcallosal Cortex                          & None            & None          \\
28 & Paracingulate Gyrus                         & None            & None          \\
29 & Anterior Cingulate Gyrus                    & None            & None          \\
30 & Posterior Cingulate Gyrus                   & Potentially     & Strong        \\
31 & Precuneous Cortex                           & Potentially     & Strong        \\
32 & Cuneal Cortex                               & None            & Potentially   \\
33 & Orbitofrontal Cortex                        & None            & Potentially   \\
34 & Parahippocampal Gyrus, Anterior Division    & None            & Strong        \\
35 & Parahippocampal Gyrus, Posterior Division   & None            & Strong        \\
36 & Lingual Gyrus                               & None            & Potentially   \\
37 & Temporal Fusiform Cortex, Anterior Division & None            & Strong        \\
38 & Temporal Fusiform Cortex, Posterior Division& None            & Strong        \\
39 & Temporooccipital Fusiform Cortex            & None            & Strong        \\
40 & Occipital Fusiform Gyrus                    & None            & Potentially   \\
41 & Frontal Operculum Cortex                    & None            & None          \\
42 & Central Opercular Cortex                    & None            & None          \\
43 & Parietal Operculum Cortex                   & None            & None          \\
44 & Planum Polare                               & None            & None          \\
45 & Heschl’s Gyrus                              & None            & None          \\
46 & Planum Temporale                            & None            & None          \\
47 & Supracalcarine Cortex                       & None            & None          \\
48 & Occipital Pole                              & None            & None          \\
\hline
\end{tabular}
\caption{\small Brain regions and their relevance with PD and AD.
The 48 cortical regions were categorized into three groups: Strong, Potentially and None.
The 48 cortical regions were categorized into three groups: Strong, Potentially, and None.
For PD, Strong denotes regions directly involved in motor control; Potentially denotes regions that show significant differences between PD patients and healthy controls and are likely involved in motor modulation; and None denotes regions with weak or insufficient evidence of association with PD.
For AD, Strong denotes regions known to undergo early and prominent neurodegeneration; Potentially denotes regions that frequently exhibit structural or functional alterations in AD but with less consistent or secondary involvement; and None denotes regions with limited or currently insufficient evidence supporting a direct association with AD-related pathology.
}
\label{tab: disease_relevance}
\end{table*}

\section{Bias Correction}
\label{sec: Bias Correction}

In brain age prediction, a common artifact is age bias: predicted age shifts with chronological age, typically overestimating younger individuals and underestimating older ones. This induces a strong correlation between the prediction error and age, undermining evaluation.
To mitigate this, we perform brain-region-level bias correction by using the training set (composed entirely of healthy controls). 
The steps can be formulated as:
\begin{align}
& \mathcal{F}_{\text{bias}} \left( y_{\text{chron}}^{(n)} \right)
= \left(\lambda_0^{(r)} + \lambda_1^{(r)} y_{\text{chron}}^{(n)} \right) - y_{\text{chron}}^{(n)}, \\
& \hat{y}_{\text{correct}}^{(r)} = \hat{y}_{\pmb{\mathbb{S}}\text{-final}}^{(r,n)} - \mathcal{F}_{\text{bias}} \left( y_{\text{chron}}^{(n)} \right),
\label{eqa: age_bias_correction}
\end{align}
where 
$\mathcal{F}_{\text{bias}}$ denotes the bias function;
$\lambda_0^{(r)}$, $\lambda_1^{(r)}$ are brain-region-specific regression coefficients fitted on healthy controls;
$\hat{y}_{\text{correct}}^{(r)}$ indicates the corrected brain age of the $r$-th region, which is the final output.

\section{Supplemented Details of Metric}

In our main text, we use the function of Maximum Mean Discrepancy (MMD). Here, we show the details of MMD.

Define the training and test distributions of brain age gaps for region $r$ as:
\begin{align}
& \mathcal{H}_\text{tr}^{(r)} =\{ \Delta\text{ReBA}^{(r,n_\text{tr})}, n_\text{tr} \in \mathcal{D}_\text{tr}^{\text{HC}}\}_{n_\text{tr}=1}^{N_\text{tr}},\\
& \mathcal{H}_\text{ts}^{(r)} =\{ \Delta\text{ReBA}^{(r,n_\text{ts})}, n_\text{ts} \in \mathcal{D}_\text{ts}^{\text{HC}}\}_{n_\text{ts}=1}^{N_\text{ts}},
\label{eqa: distributions}
\end{align}

Define the Radial Basis Function (RBF) kernel as:
\begin{align}
& k(a,b) = exp \left(- \frac{(a-b)^2}{2m^2} \right),
\label{eqa: RBF_kernel}
\end{align}
where $m$ is the median of $\mathcal{H}_\text{tr}^{(r)} \cup \mathcal{H}_\text{ts}^{(r)}$.

\begin{equation}
\begin{split}
\Phi_{\text{MMD}} 
&= 
\frac{1}{N_\text{tr}(N_{tr-1})} \sum_{n_\text{tr} \neq n_\text{tr}'} k\left(\Delta\text{ReBA}^{(r,n_\text{tr})},\Delta\text{ReBA}^{(r,n_\text{tr}')} \right) \\
&+
\frac{1}{N_\text{ts}(N_{ts-1})} \sum_{n_\text{ts} \neq n_\text{ts}'} k\left(\Delta\text{ReBA}^{(r,n_\text{ts})},\Delta\text{ReBA}^{(r,n_\text{ts}')} \right) \\
&-
\frac{2}{N_\text{tr}N_\text{ts}} \sum_{n_\text{tr}, n_\text{ts}} k\left(\Delta\text{ReBA}^{(r,n_\text{tr})},\Delta\text{ReBA}^{(r,n_\text{ts})} \right) 
\end{split}
\label{eqa: MMD}
\end{equation}

\section{Supplemented Experiments}
\label{sec: Supplemented Experiments}
The details about data are listed in Tab.~\ref{tab: dataset}.

\begin{table*}[t]
    \centering
    \label{tab:datasets}
    \begin{tabular}{c p{3.5cm} c p{10cm}} 
    \toprule
    \textbf{Group} & \textbf{Dataset Name} & \textbf{Samples} & \textbf{Access URL} \\
    \midrule
    
    \multirow{20}{*}{{\textbf{HC}}} 
      & AHDC & 120 & \url{https://openneuro.org/datasets/ds005901} \\
      & AgeRisk & 187 & \url{https://openneuro.org/datasets/ds004711} \\
      & AOMIC & 928 & \url{https://openneuro.org/datasets/ds003097} \\
      & AOMIC-PIOP1 & 226 & \url{https://openneuro.org/datasets/ds002790} \\
      & AOMIC-PIOP2 & 216 & \url{https://openneuro.org/datasets/ds002785} \\
      & BOLD-Variability & 158 & \url{https://openneuro.org/datasets/ds005270} \\
      & Chronotype-Sleep & 136 & \url{https://openneuro.org/datasets/ds003826} \\
      & DLBS & 315 & \url{https://fcon_1000.projects.nitrc.org/indi/retro/} \\
      & DP-pCASL & 186 & \url{https://openneuro.org/datasets/ds005529} \\
      & HCP & 35 & \url{https://ida.loni.usc.edu/login.jsp} \\
      & HRV-Emotion & 177 & \url{https://openneuro.org/datasets/ds003823} \\
      & IXI & 581 & \url{https://brain-development.org/ixi-dataset/} \\
      & LA5c & 272 & \url{https://openneuro.org/datasets/ds000030} \\
      & MR-ART & 148 & \url{https://openneuro.org/datasets/ds004173} \\
      & Narratives & 345 & \url{https://openneuro.org/datasets/ds002345} \\
      & NEBULA101 & 101 & \url{https://openneuro.org/datasets/ds005613} \\
      & ThinkAloud & 118 & \url{https://openneuro.org/datasets/ds006067} \\
      & TOF-MRA & 284 & \url{https://openneuro.org/datasets/ds003949} \\
      & NKI & 974 & \url{http://fcon_1000.projects.nitrc.org/indi/enhanced/} \\
      & SALD & 494 & \url{http://fcon_1000.projects.nitrc.org/indi/retro/sald.html} \\
      & SLIM & 580 & \url{http://fcon_1000.projects.nitrc.org/indi/retro/southwestuni_qiu_index.html} \\
      & GSP & 1006 & \url{https://www.neuroinfo.org/gsp} \\ 
      & \textbf{HC Total} & \textbf{7587} & \\

    \cmidrule{1-4}
    
    \multirow{2}{*}{{\textbf{PD}}} 
      & PPMI & 169 & \url{https://www.ppmi-info.org/} \\
      & In-house collection & 157 & - \\ 
      & \textbf{PD Total} & \textbf{326} & \\
    \cmidrule{1-4}
    \textbf{AD} 
      & ADNI & 107 & \url{https://adni.loni.usc.edu/data-samples/adni-data/} \\
    & \textbf{AD Total} & \textbf{107} & \\
    
    \bottomrule
    \end{tabular}
    \caption{Summary of Datasets by Group} 
    \label{tab: dataset}
\end{table*}

\subsection{Comparison on HCS Metric}

For the HCS metric, the main text reports results for only five representative brain regions. The Harvard–Oxford atlas used in our study, however, contains 48 regions in total. Here, we provide the complete set of HCS values for all regions: Regions 1–25 are shown in Fig.~\ref{fig: HCS_all_regions_1-25}, while Regions 26–48 and the overall average are presented in Fig.~\ref{fig: HCS_all_regions_26-48}.

\subsection{Comparison on NDC Metric}

The NDC metric incorporates clinical priors by specifying, for each disease, a set of regions expected to exhibit accelerated aging. It then evaluates whether these regions indeed show higher ReBA in affected individuals.
In the main text, we presented results using PD as an example. Here, we report the corresponding analysis for AD in Fig.~\ref{fig: Experiments_AD_NDC}. Clinically, AD is known to involve accelerated degeneration in regions 8–16, 20–23, 30, 31, 34, 35, 37–39. Based on these priors, we compute the NDC score by averaging the regional brain-age gaps over this set of AD-related regions and compare the resulting distributions across AD, HC, and PD subjects. The NDC for AD reaches approximately 0.71, which is substantially higher than that of HC (~0.3) and PD (~0.4), providing additional evidence supporting the validity of our regional brain-age estimation

\subsection{Training Loss Curve}

Our framework consists of two components, Teacher and Student. In Fig.~\ref{fig: training_curve}, we plot the training losses for both stages. The curves for the teacher and student losses decrease steadily without oscillation, indicating that the optimization is stable and both models converge properly. This suggests that our training strategy for the teacher–student framework is effective and does not suffer from obvious optimization difficulties or collapse.

\subsection{Sensitivity Analysis of Hyperparameters}

In this study, we consider three hyperparameters that noticeably influence model performance. In the main experiments, we set them to $\alpha = 1$, $\eta = 0.1$, and $\zeta = 1$. To examine how each parameter affects the results, we conduct a one-factor analysis where we vary a single parameter while keeping the other two fixed. See Fig.~\ref{fig: hyperparameters}.

For NDC (top row), the model is clearly sensitive to all three hyperparameters. Changing $\alpha$ within $[0,1.5]$ leads to a marked rise and subsequent drop in NDC, and varying $\eta$ induces an even stronger effect, with NDC decreasing sharply when $\eta$ exceeds a small range around $0.1$. The NDC score also increases and then declines as $\zeta$ grows, indicating a relatively narrow high-performing interval. These trends suggest that the choice of $\alpha$, $\eta$, and $\zeta$ must be made carefully when optimizing NDC.

For HCS (bottom row), the curves are comparatively flatter. Although there is a mild peak around the default setting ($\alpha = 1$, $\eta = 0.1$, $\zeta = 1$), the fluctuations in HCS across the explored ranges are much smaller than those observed for NDC. This indicates that our method is relatively robust with respect to these hyperparameters when evaluated by HCS, while NDC is more sensitive and thus more dependent on precise hyperparameter tuning.

\bibliography{example_paper}

@article{bethlehem2022brain,
  title={Brain charts for the human lifespan},
  author={Bethlehem, Richard AI and Seidlitz, Jakob and White, Simon R and Vogel, Jacob W and Anderson, Kevin M and Adamson, Chris and Adler, Sophie and Alexopoulos, George S and Anagnostou, Evdokia and Areces-Gonzalez, Ariosky and others},
  journal={Nature},
  volume={604},
  number={7906},
  pages={525--533},
  year={2022},
  publisher={Nature Publishing Group UK London}
}

@article{kalc2024brainage,
  title={BrainAGE: Revisited and reframed machine learning workflow},
  author={Kalc, Polona and Dahnke, Robert and Hoffstaedter, Felix and Gaser, Christian and Alzheimer's Disease Neuroimaging Initiative},
  journal={Human Brain Mapping},
  year={2024}
}

@article{jenkinson2012fsl,
  title={Fsl},
  author={Jenkinson, Mark and Beckmann, Christian F and Behrens, Timothy EJ and Woolrich, Mark W and Smith, Stephen M},
  journal={Neuroimage},
  volume={62},
  number={2},
  pages={782--790},
  year={2012},
  publisher={Elsevier}
}

@article{avants2009advanced,
  title={Advanced normalization tools (ANTS)},
  author={Avants, Brian B and Tustison, Nick and Song, Gang and others},
  journal={Insight j},
  volume={2},
  number={365},
  pages={1--35},
  year={2009}
}

@article{wang2025full,
  title={A full life cycle biological clock based on routine clinical data and its impact in health and diseases},
  author={Wang, Kai and Liu, Fei and Wu, Wei and Hu, Changxi and Shen, Xian and Wang, Meihao and Li, Gen and Zeng, Fanxin and Liu, Li and Wong, Io Nam and others},
  journal={Nature Medicine},
  pages={1--11},
  year={2025},
  publisher={Nature Publishing Group US New York}
}

@article{marek2011parkinson,
  title={The Parkinson progression marker initiative (PPMI)},
  author={Marek, Kenneth and Jennings, Danna and Lasch, Shirley and Siderowf, Andrew and Tanner, Caroline and Simuni, Tanya and Coffey, Chris and Kieburtz, Karl and Flagg, Emily and Chowdhury, Sohini and others},
  journal={Progress in Neurobiology},
  volume={95},
  number={4},
  pages={629--635},
  year={2011},
  publisher={Elsevier}
}

@article{mueller2005alzheimer,
  title={The Alzheimer's disease neuroimaging initiative},
  author={Mueller, Susanne G and Weiner, Michael W and Thal, Leon J and Petersen, Ronald C and Jack, Clifford and Jagust, William and Trojanowski, John Q and Toga, Arthur W and Beckett, Laurel},
  journal={Neuroimaging Clinics},
  volume={15},
  number={4},
  pages={869--877},
  year={2005},
  publisher={Elsevier}
}

@article{lee2022deep,
  title={Deep learning-based brain age prediction in normal aging and dementia},
  author={Lee, Jeyeon and Burkett, Brian J and Min, Hoon-Ki and Senjem, Matthew L and Lundt, Emily S and Botha, Hugo and Graff-Radford, Jonathan and Barnard, Leland R and Gunter, Jeffrey L and Schwarz, Christopher G and others},
  journal={Nature Aging},
  volume={2},
  number={5},
  pages={412--424},
  year={2022},
  publisher={Nature Publishing Group US New York}
}

@inproceedings{jang2022m3t,
  title={M3t: three-dimensional medical image classifier using multi-plane and multi-slice transformer},
  author={Jang, Jinseong and Hwang, Dosik},
  booktitle={Proceedings of the IEEE/CVF Computer Vision and Pattern Recognition Conference},
  pages={20718--20729},
  year={2022}
}

@inproceedings{tang2022self,
  title={Self-supervised pre-training of swin transformers for 3d medical image analysis},
  author={Tang, Yucheng and Yang, Dong and Li, Wenqi and Roth, Holger R and Landman, Bennett and Xu, Daguang and Nath, Vishwesh and Hatamizadeh, Ali},
  booktitle={Proceedings of the IEEE/CVF Computer Vision and Pattern Recognition Conference},
  pages={20730--20740},
  year={2022}
}

@inproceedings{wald2025revisiting,
  title={Revisiting MAE pre-training for 3D medical image segmentation},
  author={Wald, Tassilo and Ulrich, Constantin and Lukyanenko, Stanislav and Goncharov, Andrei and Paderno, Alberto and Miller, Maximilian and Maerkisch, Leander and Jaeger, Paul and Maier-Hein, Klaus},
  booktitle={Proceedings of the IEEE/CVF Computer Vision and Pattern Recognition Conference},
  pages={5186--5196},
  year={2025}
}

@inproceedings{chen2023masked,
  title={Masked image modeling advances 3d medical image analysis},
  author={Chen, Zekai and Agarwal, Devansh and Aggarwal, Kshitij and Safta, Wiem and Balan, Mariann Micsinai and Brown, Kevin},
  booktitle={Proceedings of the IEEE/CVF Winter Conference on Applications of Computer Vision},
  pages={1970--1980},
  year={2023}
}

@inproceedings{zhao2023ae,
  title={AE-FLOW: Autoencoders with normalizing flows for medical images anomaly detection},
  author={Zhao, Yuzhong and Ding, Qiaoqiao and Zhang, Xiaoqun},
  booktitle={The Eleventh International Conference on Learning Representations},
  year={2023}
}

@article{ashburner2000voxel,
  title={Voxel-based morphometry—the methods},
  author={Ashburner, John and Friston, Karl J},
  journal={NeuroImage},
  volume={11},
  number={6},
  pages={805--821},
  year={2000},
  publisher={Elsevier}
}

@article{good2001voxel,
  title={A voxel-based morphometric study of ageing in 465 normal adult human brains},
  author={Good, Catriona D and Johnsrude, Ingrid S and Ashburner, John and Henson, Richard NA and Friston, Karl J and Frackowiak, Richard SJ},
  journal={NeuroImage},
  volume={14},
  number={1},
  pages={21--36},
  year={2001},
  publisher={Elsevier}
}

@article{pennanen2005voxel,
  title={A voxel based morphometry study on mild cognitive impairment},
  author={Pennanen, C and Testa, C and Laakso, MP and Hallikainen, M and Helkala, EL and H{\"a}nninen, T and Kivipelto, M and K{\"o}n{\"o}nen, M and Nissinen, A and Tervo, S and others},
  journal={Journal of Neurology, Neurosurgery \& Psychiatry},
  volume={76},
  number={1},
  pages={11--14},
  year={2005},
  publisher={BMJ Publishing Group Ltd}
}

@article{seitz2024brainage,
  title={BrainAGE, brain health, and mental disorders: A systematic review},
  author={Seitz-Holland, Johanna and Haas, Shalaila S and Penzel, Nora and Reichenberg, Abraham and Pasternak, Ofer},
  journal={Neuroscience \& Biobehavioral Reviews},
  volume={159},
  pages={105581},
  year={2024},
  publisher={Elsevier}
}

@article{yu2024brain,
  title={Brain-age prediction: Systematic evaluation of site effects, and sample age range and size},
  author={Yu, Yuetong and Cui, Hao-Qi and Haas, Shalaila S and New, Faye and Sanford, Nicole and Yu, Kevin and Zhan, Denghuang and Yang, Guoyuan and Gao, Jia-Hong and Wei, Dongtao and others},
  journal={Human Brain Mapping},
  volume={45},
  number={10},
  pages={e26768},
  year={2024},
  publisher={Wiley Online Library}
}

@article{kuchcinski2023mri,
  title={MRI BrainAGE demonstrates increased brain aging in systemic lupus erythematosus patients},
  author={Kuchcinski, Gr{\'e}gory and Rumetshofer, Theodor and Zervides, Kristoffer A and Lopes, Renaud and Gautherot, Morgan and Pruvo, Jean-Pierre and Bengtsson, Anders A and Hansson, Oskar and J{\"o}nsen, Andreas and Sundgren, Pia C Maly},
  journal={Frontiers in Aging Neuroscience},
  volume={15},
  pages={1274061},
  year={2023},
  publisher={Frontiers Media SA}
}

@article{jonsson2019brain,
  title={Brain age prediction using deep learning uncovers associated sequence variants},
  author={J{\'o}nsson, Benedikt Atli and Bjornsdottir, Gyda and Thorgeirsson, Thorgeir E and Ellingsen, Lotta Mar{\'\i}a and Walters, G Bragi and Gudbjartsson, Daniel Fannar and Stefansson, Hreinn and Stefansson, Kari and Ulfarsson, Magnus Orn},
  journal={Nature communications},
  volume={10},
  number={1},
  pages={5409},
  year={2019},
  publisher={Nature Publishing Group UK London}
}

@article{armanious2021age,
  title={Age-Net: an MRI-based iterative framework for brain biological age estimation},
  author={Armanious, Karim and Abdulatif, Sherif and Shi, Wenbin and Salian, Shashank and K{\"u}stner, Thomas and Weiskopf, Daniel and Hepp, Tobias and Gatidis, Sergios and Yang, Bin},
  journal={IEEE Transactions on Medical Imaging},
  volume={40},
  number={7},
  pages={1778--1791},
  year={2021},
  publisher={IEEE}
}

@article{cheng2021brain,
  title={Brain age estimation from MRI using cascade networks with ranking loss},
  author={Cheng, Jian and Liu, Ziyang and Guan, Hao and Wu, Zhenzhou and Zhu, Haogang and Jiang, Jiyang and Wen, Wei and Tao, Dacheng and Liu, Tao},
  journal={IEEE Transactions on Medical Imaging},
  volume={40},
  number={12},
  pages={3400--3412},
  year={2021},
  publisher={IEEE}
}

@article{cole2015prediction,
  title={Prediction of brain age suggests accelerated atrophy after traumatic brain injury},
  author={Cole, James H and Leech, Robert and Sharp, David J and Alzheimer's Disease Neuroimaging Initiative},
  journal={Annals of Neurology},
  volume={77},
  number={4},
  pages={571--581},
  year={2015},
  publisher={Wiley Online Library}
}

@article{wang2011mri,
  title={MRI-based age prediction using hidden Markov models},
  author={Wang, Bing and Pham, Tuan D},
  journal={Journal of Neuroscience Methods},
  volume={199},
  number={1},
  pages={140--145},
  year={2011},
  publisher={Elsevier}
}

@article{liem2017predicting,
  title={Predicting brain-age from multimodal imaging data captures cognitive impairment},
  author={Liem, Franziskus and Varoquaux, Ga{\"e}l and Kynast, Jana and Beyer, Frauke and Masouleh, Shahrzad Kharabian and Huntenburg, Julia M and Lampe, Leonie and Rahim, Mehdi and Abraham, Alexandre and Craddock, R Cameron and others},
  journal={NeuroImage},
  volume={148},
  pages={179--188},
  year={2017},
  publisher={Elsevier}
}

@article{sarasso2021progression,
  title={Progression of grey and white matter brain damage in Parkinson's disease: a critical review of structural MRI literature},
  author={Sarasso, Elisabetta and Agosta, Federica and Piramide, Noemi and Filippi, Massimo},
  journal={Journal of Neurology},
  volume={268},
  number={9},
  pages={3144--3179},
  year={2021},
  publisher={Springer}
}

@article{liu2020brain,
  title={Brain functional and structural signatures in Parkinson’s disease},
  author={Liu, Chunhua and Jiang, Jiehui and Zhou, Hucheng and Zhang, Huiwei and Wang, Min and Jiang, Juanjuan and Wu, Ping and Ge, Jingjie and Wang, Jian and Ma, Yilong and others},
  journal={Frontiers in Aging Neuroscience},
  volume={12},
  pages={125},
  year={2020},
  publisher={Frontiers Media SA}
}

@article{kaufmann2019common,
  title={Common brain disorders are associated with heritable patterns of apparent aging of the brain},
  author={Kaufmann, Tobias and van der Meer, Dennis and Doan, Nhat Trung and Schwarz, Emanuel and Lund, Martina J and Agartz, Ingrid and Aln{\ae}s, Dag and Barch, Deanna M and Baur-Streubel, Ramona and Bertolino, Alessandro and others},
  journal={Nature Neuroscience},
  volume={22},
  number={10},
  pages={1617--1623},
  year={2019},
  publisher={Nature Publishing Group US New York}
}

@article{taylor2022investigating,
  title={Investigating the temporal pattern of neuroimaging-based brain age estimation as a biomarker for Alzheimer's Disease related neurodegeneration},
  author={Taylor, Alexei and Zhang, Fengqing and Niu, Xin and Heywood, Ashley and Stocks, Jane and Feng, Gangyi and Popuri, Karteek and Beg, Mirza Faisal and Wang, Lei and Alzheimer's Disease Neuroimaging Initiative and others},
  journal={Neuroimage},
  volume={263},
  pages={119621},
  year={2022},
  publisher={Elsevier}
}

@article{lee2022regional,
  title={Regional rather than global brain age mediates cognitive function in cerebral small vessel disease},
  author={Lee, Pei-Lin and Kuo, Chen-Yuan and Wang, Pei-Ning and Chen, Liang-Kung and Lin, Ching-Po and Chou, Kun-Hsien and Chung, Chih-Ping},
  journal={Brain Communications},
  volume={4},
  number={5},
  pages={fcac233},
  year={2022},
  publisher={Oxford University Press US}
}

@article{busby2024regional,
  title={Regional brain aging: premature aging of the domain general system predicts aphasia severity},
  author={Busby, Natalie and Newman-Norlund, Sarah and Sayers, Sara and Rorden, Chris and Newman-Norlund, Roger and Wilmskoetter, Janina and Roth, Rebecca and Wilson, Sarah and Schwen-Blackett, Deena and Kristinsson, Sigfus and others},
  journal={Communications Biology},
  volume={7},
  number={1},
  pages={718},
  year={2024},
  publisher={Nature Publishing Group UK London}
}

@article{wu2025regional,
  title={Regional Brain Aging Disparity Index: Region-Specific Brain Aging State Index for Neurodegenerative Diseases and Chronic Disease Specificity},
  author={Wu, Yutong and Sun, Shen and Zhang, Chen and Ma, Xiangge and Zhu, Xinyu and Li, Yanxue and Lin, Lan and Fu, Zhenrong},
  journal={Bioengineering},
  volume={12},
  number={6},
  pages={607},
  year={2025},
  publisher={MDPI}
}

@article{burciu2018imaging,
  title={Imaging of motor cortex physiology in Parkinson's disease},
  author={Burciu, Roxana G and Vaillancourt, David E},
  journal={Movement Disorders},
  volume={33},
  number={11},
  pages={1688--1699},
  year={2018},
  publisher={Wiley Online Library}
}

@article{gao2016study,
  title={The study of brain functional connectivity in Parkinson’s disease},
  author={Gao, Lin-lin and Wu, Tao},
  journal={Translational Neurodegeneration},
  volume={5},
  number={1},
  pages={18},
  year={2016},
  publisher={Springer}
}

@article{riccardi2025distinct,
  title={Distinct brain age gradients across the adult lifespan reflect diverse neurobiological hierarchies},
  author={Riccardi, Nicholas and Teghipco, Alex and Newman-Norlund, Sarah and Newman-Norlund, Roger and Rangus, Ida and Rorden, Chris and Fridriksson, Julius and Bonilha, Leonardo},
  journal={Communications Biology},
  volume={8},
  number={1},
  pages={802},
  year={2025},
  publisher={Nature Publishing Group UK London}
}

@article{ren2025deepprep,
  title={DeepPrep: an accelerated, scalable and robust pipeline for neuroimaging preprocessing empowered by deep learning},
  author={Ren, Jianxun and An, Ning and Lin, Cong and Zhang, Youjia and Sun, Zhenyu and Zhang, Wei and Li, Shiyi and Guo, Ning and Cui, Weigang and Hu, Qingyu and others},
  journal={Nature Methods},
  pages={1--4},
  year={2025},
  publisher={Nature Publishing Group US New York}
}

@article{fischl2012freesurfer,
  title={FreeSurfer},
  author={Fischl, Bruce},
  journal={Neuroimage},
  volume={62},
  number={2},
  pages={774--781},
  year={2012},
  publisher={Elsevier}
}
\bibliographystyle{icml2026}




\end{document}